# ValiText - a Unified Validation Framework for Computational Text-Based Measures of Social Constructs

***Lukas Birkenmaier[1], Clemens Lechner[1], Claudia Wagner[1,2,3]**

*[1]GESIS – Leibniz Institute for the Social Sciences, Mannheim, Germany*

*[2]RWTH Aachen University, Germany*

*[3]Complexity Science Hub Vienna, Austria*

*KEYWORDS: Text as Data, Validity, Measurement, Social Science Constructs, Systematic Review, Qualitative Expert Interviews*

**ABSTRACT:** Guidance on how to validate computational text-based measures of social constructs is fragmented. While researchers generally acknowledge the importance of validating text-based measures, they often lack a shared vocabulary and a unified framework to do so. This paper introduces ValiText, a new validation framework designed to assist scholars in validly measuring social constructs in textual da-ta. The framework is built on a conceptual foundation of validity in the social sciences, strengthened by an empirical review of validation practices in the social sciences and consultations with experts. Ultimately, ValiText prescribes researchers to demonstrate three types of validation evidence: substantive evidence (outlining the theoretical underpinning of the measure), structural evidence (examining the properties of the text model and its output) and external evidence (testing for how the measure relates to independent information). The framework is further supplemented by a checklist of validation steps, offering practical guidance in the form of documentation sheets that guide researchers in the validation process.

*Corresponding Author: lukas.birkenmaier@gesis.org

**Link to the Shiny App: https://www.ValiTex.info/**

## 1. Introduction

Computational text-based measures of social constructs are difficult to validate (Baden et al., 2020; Kantner & Overbeck, 2020). However, validation is necessary because measures derived from textual data are often used as variables in downstream analysis. Hence, if a text-based measure inaccurately represents the concept or phenomenon it aims to capture, this can introduce systematic bias and errors in the research process. Therefore, in order to determine whether a measure is a valid operationalization of the underlying construct, it is crucial to thoroughly validate the relationship between text-based measures and theoretical constructs (Hughes, 2018). To do so, the most established method is comparing the output scores of computational methods with human annotations that serve as a gold-standard benchmark (Grimmer et al., 2022).

In this paper, we draw attention to critical opinions suggesting that the mere comparison of output scores with human annotations is insufficient for thorough validation. This view is supported by a long-standing body of research in the social sciences, which emphasizes the need to present multiple forms of validation evidence to validate measures of social constructs effectively (Flake et al., 2017). In computational text analysis, social scientists routinely employ a broad spectrum of validation strategies going beyond simple human annotation comparison (Birkenmaier et al., 2023). However, strategies for validation are often diffuse, or lack justification altogether. For instance, methodological frameworks and best-practices are often transferred from specific social science subfields without fully accounting for the specific nature of text-based measures (e.g., Adcock & Collier, 2001). Additionally, inconsistencies in terminology further complicate the matter, as variability in language and definitions can cause misunderstandings, or lead to validation steps being deliberatively left out in the reporting of results (Birkenmaier et al., 2023).

To address these concerns, this paper introduces *ValiText* - a unified validation framework for computational text-based measures of social constructs. As such, the framework aims to fulfill three objectives: a) to stress the importance for researchers to provide different kinds of validation evidence for computational text analysis, b) to define the key types of validation evidence that are required for sufficient validation, c) to guide researchers which concrete validation steps to apply for their text-based research, and d) to provide a documentation template in the form of a checklist that can be used to document validation efforts effectively.

## 2. Bias and Error in Computational Text Analysis

Validation is a necessary requirement for computational text analysis (Grimmer et al., 2022; Krippendorff, 2018). Essentially, the purpose of validation is to uncover and correct mistakes in the measurement process, ensuring that the output scores are an accurate reflection of the true nature of the construct being studied. Mistakes in measurement can arise from different sources, with varying consequences for the validity of the measurement.

On the one hand, random mistakes (*errors*) can be seen as uncontrolled and random errors that occur due to the limitations of the research design. These errors are not inherently problematic for the measurement because it's assumed they will neutralize each other. For text analysis, errors can be attributed to the inherent complex and multi-dimensional nature of text, as well as to the uncertainty about how social constructs are depicted in texts. For instance, language can often be interpreted in different ways depending on the readers perspective, knowledge, or cultural background. Examples of these characteristics are the presence of sarcasm or irony (Ravi & Ravi, 2015), ambiguity and polysemy (Boxman-Shabtai, 2020; Roberts, 1989), or context-specific references and interpretations (Krippendorff, 2018; Mayring, 2004). Furthermore, the construct that is to be measured (i.e., an emotion or a political topic) can often be expressed differently in text, thus allowing for multiple ways to operationalize the construct of interest (Drisko & Maschi, 2016; Krippendorff, 2018).

On the other hand, systematic errors (*biases*) are more problematic, as biases can severely compromise the validity of measurements. This is because social constructs are typically used for further analyses as dependent or independent variables to draw insights about real-world relationships. Biases arising from incorrect measurements of constructs can negatively affect the clarity of these relationships, thus severely affecting subsequent analytical processes. For text analysis, such biases may result from methodological flaws or inherent prejudices in the methods or strategies applied by researchers. For example, large language models (LLMs) may introduce bias in the measurement process via the data used to train and test the models, either via data leakage (Gibney, 2022; Kapoor & Narayanan, 2022), non-representative datasets in respect to the measured construct (Cai et al., 2022), or model-inherent human stereotypes (van Giffen et al., 2022).

Likewise, the text analysis pipeline is usually characterized by high researchers' degrees of freedom (i.e., the various choices of collecting and analyzing data, see Bakker et al. (2020)). Because computational methods require a variety of analytical choices, starting from construct operationalization, model selection, data pre-processing and model hyperparameters, researchers' decisions often affect the outcomes of analyses significantly (Arnold et al., 2023).

Ultimately, validation enables the researcher to identify bias – thereby reducing the uncertainty around the measurement process. In other words, without sufficient validation, researchers can never completely rely on the accuracy of their empirical findings, regardless of whether they turn out to be correct or inaccurate.

## 3. A Critical Perspective on Human Annotations as a Gold-Standard for Text Analysis

The most common way to ensure valid measurements is comparing text-based measures with human annotations, often referred to as "gold-standard" data (Grimmer et al., 2022). These human annotated labels are assumed to be entirely accurate and objective to provide a sufficient benchmark for the true underlying concepts. The foundation of this approach rests on the belief that human comprehension of texts surpasses that of text models and that, when adequately trained, humans can achieve the most accurate and valid text judgments (Song et al., 2020). Although it's widely accepted that human annotations are essential to demonstrate validity, there is significant critique challenging the notion that human annotations alone are sufficient for validation.

First and foremost, we must constantly be reminded that objective "gold-standard" data does not exist. For instance, the content analysis literature, which focuses on manually describing communication data using trained coders, usually neglects the notion of a true underlying “gold standard” labels. Despite meticulous efforts to ensure objectivity- such as providing comprehensive construct definitions, utilizing double-blind coding, conducting inter-coder reliability assessments, and thoroughly training coders (Krippendorff, 2018) - the literature emphasizes the inherent variability in human perception. The differing interpretations suggest that the context in which a text's content is perceived plays a crucial role in determining its meaning (Drisko & Maschi, 2016). Thus, we need to detach ourselves from the idea that gold-standard targets in the form of objective truths exist after all for social constructs in texts.

Building upon this observation, human annotations frequently fall short of acceptable quality standards. In particular when annotations are sourced from crowd workers or untrained annotators, the quality of annotations can diminish due to factors like varied levels of expertise, differing interpretation of guidelines, and potential lack of motivation (Chmielewski & Kucker, 2020).

Relying on Monte-Carlo Simulation and simulated data, Song et al. (2020) vividly illustrate that low-quality human annotations can impact the results derived from empirical measures. In their study, Song et al. systematically varied factors such as the sampling variability, the number of human coders, the size of the validation data, or the required level of intercoder reliability, and showed that these factors significantly affected the classification accuracy on the “true” labels of their simulated textual data.

Likewise, when researchers rely on surface-level correlations, it often prevents them from gaining context-specific insights into the measurement. This can lead to flawed conclusions, potentially caused by systematic bias or low quality data (see Daikeler et al., 2024). A striking illustration of this is provided by Hirst et al. (2014), who showed that a machine classifier designed to identify political ideologies erroneously focused on incumbency. This error occurred because incumbency was more easily detected by the model and coincidentally correlated with ideology in their dataset. To identify these more fine-grained limitations, more nuanced validation strategies beyond output comparison are required.

## 4. Deriving the Validation Framework

We suggest a validation framework for text analysis that complements the mere comparison of output scores with human annotations. We build upon earlier work of a systematic review conducted in the field of political communication (as outlined in Birkenmaier et al. (2023)). In the systematic review, we documented more than 20 different types of validation evidence and grouped them into overarching categories. Furthermore, we conducted expert interviews to further document additional activities that researchers undertake during validation, but do not usually report.

Ultimately, the development of the ValiText framework followed two main stages: first, we defined the fundamental types of validation evidence needed for text-based measures of social constructs. To do so, we extended the broad categories of validation evidence identified in the systematic review and embed them into well-established practices within the social science literature. Second, we gather an extensive list of empirical validation steps (i.e., specific tests that can be executed to produce validation evidence) for each type of validation evidence. Each step is meticulously detailed in a checklist, providing practical guidance to researchers for different use cases.

### *Types of Validation Evidence*

There are multiple approaches on how to separate different dimensions of validation evidence for text analysis conceptually (Fang et al., 2022; Grimmer et al., 2022; Quinn et al., 2010).[1] We initially build on the conceptualization introduced in Birkenmaier et al. (2023) who differentiate between *internal* and *external* validation evidence. Whereas internal validation relates to the systematic assessment of the measurement model and its output, external validation relates to the comparison of output scores with independent information. Empirically, these two fundamental types of validation evidence serve as a useful distinction of the underlying motivations in the validation of text-based measures, i.e., understanding the characteristics and limitations of the measurement process (internal validation) and demonstrating that the output scores are empirically linked to alternative measures or criteria (external validation).

[1] For a detailed discussion, see Birkenmaier et al. (2023)

In order to integrate our framework more deeply with established measurement theory, we build upon and expand Loevinger's (1957) conceptual validation framework for use in computational text analysis. At its core, the conceptualization of Loevinger (1975) incorporates three fundamental types of validation evidence, i.e., *substantive, structural*, and *external* evidence. Traditionally, these types of evidence are associated with the developmental phases of creating measurement instruments for social constructs: constitution of the item pool based on theory and construct definition, analysis and selection of the internal item structure, and correlations of the construct with external constructs (Clark & Watson, 2019). As such, these three types of evidence are mutually exclusive and exhaustive to all types of validation evidence, and mandatory to demonstrate measurement validity. Figure 1 presents these fundamental types of validation evidence adapted to the context of text analysis. As described in more detail below, *substantive* and *structural* evidence pertain to internal validation, and *external* evidence pertains to external validation. For text analysis, to provide *substantive evidence*, researchers need to outline the theoretical underpinning of the measurement. To provide *structural evidence,* researchers should examine and evaluate the properties of the text model and its output. To provide *external evidence*, researchers should test for how the empirical measures relate to other independent information or criteria.

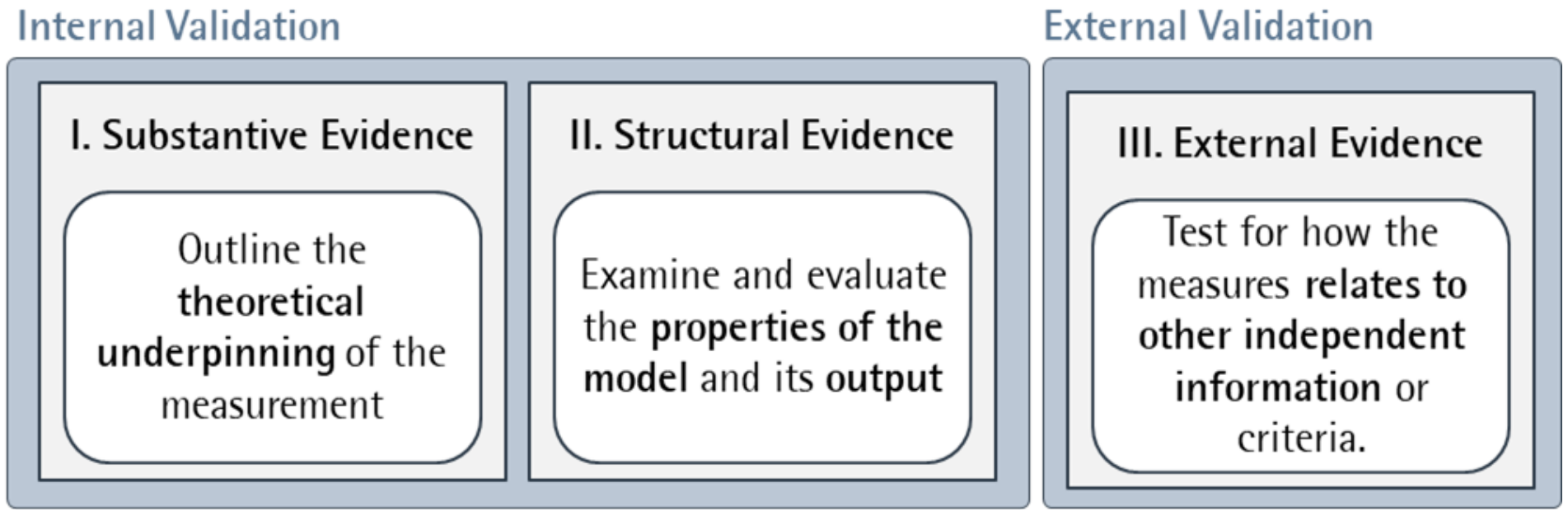

*Figure 1: Three Fundamental Types of Validation Evidence*

One advantage of adapting this tripartite structure of substantive, structural, and external evidence is to avoid confusion with established, but often confusingly used validity terms such as "discriminant validity" or "face validity" in the context of text analysis. Nonetheless, we map established validity terms to each stage of the framework and provide a complete overview in Appendix 1.

Lastly, we also add a separate section about robustness checks (i.e., reporting alternative specifications of the measurement). As Grimmer et al. (2023) note, small changes across the measurement design can dramatically change measurement outcomes, and a transparent reporting of alternative measurement designs effectively provides evidence for measurement validity.

*Selecting Validation Steps*

For the selection of empirical validation steps to the checklist, we initially start by the documentation of validation steps identified in Birkenmaier et al. (2023). For instance, examples for internal validation are *feature importance analysis,* and *comparison with human annotations* for external validation. In addition, we also consider validation steps that were brought up in the expert interviews with domain experts as validation activities that are rarely reported, such as *manual pre-coding of human annotators* or *qualitative strategies for error analysis*.

One aspect to overcome is that some validation steps only apply to a specific type of text-based method, such as inspection of *topic coherence metrics* for topic modelling, or require different techniques for different methods, such as *feature importance analysis*. Therefore, we initially define the validation steps in a broad way to capture the underlying idea of each validation step. In the checklists, however, we provide a more detailed overview of different use cases, also including potentially context-specific, tough meaningful validation steps. To identify validation steps beyond the ones reported in social science publications, one of the authors with a background in computer science further complemented the existing list of validation steps with an evaluated opinion on state-of-the-art steps validation strategies in the broader research field of natural language processing (NLP).

After selecting the validation steps, we carefully classify and document each validation step in the checklist. For each validation step, we include its name, a brief description, implementation methods, and a classification indicating whether the validation step is generally applicable or context-dependent. Additionally, we provide references to practical applications and further literature.

Nonetheless, it is important to note that our current selection and description of validation steps does not claim to be complete, nor that new strategies and tools should not be included. Therefore, we implement a participatory process that will allow researchers to add and discuss specific validation steps via an open repository on GitHub that allows researchers to subsequently suggest new validation steps and practical tools that can be documented using the frameworks' guiding structure.

## 5. Overview Framework

Figure 2 presents the final framework distinguishing between different subtypes of validation evidence. In the following chapters, we will primarily discuss the core subtypes of validation evidence and the respective validation steps in the checklist in greater detail. In Appendix 2 we further discuss the context-specific subtypes of validation evidence that can further supplement validation efforts.

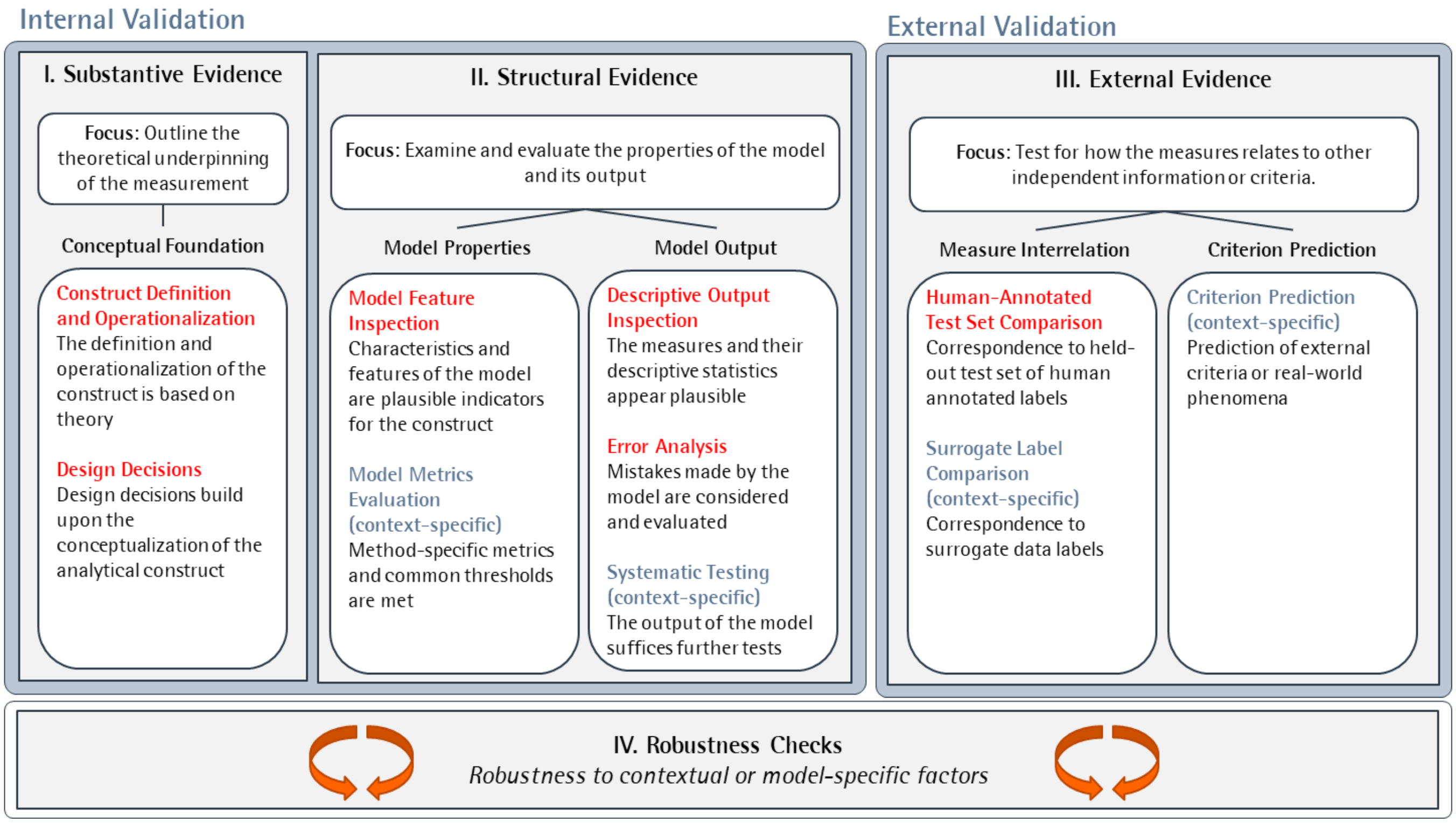

*Figure 2: ValiText Framework*

*I. Substantive Evidence*

Before conducting any measurements, researchers should outline the theoretical underpinning of the measurement to demonstrate *substantive evidence*. This can be done by documenting and providing justification for all conceptual considerations prior to measurement. Validation steps for substantive evidence should therefore demonstrate that the measurement is based on a strong conceptual foundation and that all design decisions are sufficiently justified based on the conceptualization of the construct.

Table 1 displays the validation steps that provide substantive evidence.

*Table 1: Validation Steps for Substantive Evidence (context-independent)*

| ID | Validation Steps | Considerations | Implementation | Reference |
|---|---|---|---|---|
| | | **Conceptual Foundation** | | |
| **Construct Definition & Operationalization** | | | | |
| **I.1** | **Documentation of the conceptual background** | Have I conducted a literature review or consulted with domain experts to gain a sufficient understanding of conceptual background of the construct? | Summarizing existing literature on the conceptual background of the construct | Krippendorf (2018) |
| **I.2** | **Documentation of the operationalization and codebook development** | Have I sufficiently explained how the construct should manifest itself in the textual data? Have I documented my operationalization in a codebook? | Providing definition and conceptualization of the construct | Krippendorf (2018) |
| **I.3** | **Manual Pre-coding** | Have I reached sufficient interrater agreement for a subsample of the textual data? Have I ensured that the construct can be detected in the textual data? Have I outlined my rules of coding uncertainty across coders? | Reaching sufficient interrater agreement (e.g., Krippendorff's alpha α) | Krippendorf (2018), Plank (2022) |
| **Design Decisions** | | | | |
| **I.4** | **Justification of data collection decisions** | Have I selected a dataset that is representative and relevant to the research question and population of interest? Have I justified the data selection decisions (e.g., using keywords)? Have I assessed the quality and completeness of the dataset and checked for potential biases or inconsistencies? | Outlining the rationale behind data selection / collection decisions; Documenting potential limitations and data quality issues | Krippendorf (2018) |
| **I.5** | **Justification of method choice** | Have I selected the appropriate type of method based on the operationalization of the construct and data characteristics? Have I justified the concrete selection of a particular model? | Outlining the rationale behind method selection; Documenting potential limitations in comparison to alternative methods | Grimmer et al. (2022) |
| **I.6** | **Justification of the level of analysis** | Have I selected the appropriate level of analysis? Have I considered potential problems when aggregating measures from lower to higher levels (e.g., sentence to paragraph level)? | Outline the rationale behind the selected level of analysis (e.g., token, sentence, or paragraph level). | Jankowski & Huber (2022) |
| **I.7** | **Justification of preprocessing decisions** | Have I justified relevant changes to the text prior to the analysis, such as removing certain words or phrases? | Outlining the rationale behind preprocessing decisions | Grimmer et al. (2022) |

*Note: An interactive version of the checklist is available online at https://www.valitex.info/*

***Construct definition and operationalization.*** Usually, researchers should start with *documenting the conceptual background* (I.1) of the construct. This may involve referencing previous definitions or research on the construct's dimensionality and its manifestation in language. Next, researchers should provide a *justification for the operationalization* (I.2), creating a connection between the construct definition and the textual data. As an example, if researchers aim to measure polarization in political discourse, they could make the argument that heightened polarization is likely to result in the use of more distinct vocabulary, as supported by previous studies (Lowe & Benoit, 2013; Peterson & Spirling, 2018). To test their operationalization, researchers then need to conduct *manual precoding* (I.3) where two or more coders manually label a subset of the data to determine if the construct is sufficiently defined and discernible by human coders.[2] If researchers see that the coding task is too subjective, they can adjust their operationalization, possibly considering "learning from disagreement" methods (see Uma et al., 2021).

***Design Decisions.*** Based on the construct definition, researchers must demonstrate that their design decisions are rooted in the conceptualization of the analytical construct. Crucial design decisions involve outlining the criteria for data collection (I.4), method selection (I.5), level of analysis (I.6) and preprocessing decisions (I.7). Broadly construed, design decisions should primarily be derived from the construct definition and its operationalization. This is to navigate through what Gelman and Loken (2016) call a "garden of forking paths" – having to take and justify multiple decisions in a setting of high researchers degree of freedom.

[2] If human coders are not able to correctly identify a construct, any computational method is likely to fail as well, possibly by taking up spurious relations in the data not connected to the construct of interest.

For instance, for the *justification of data collection decisions* (I.4), researchers can critically discuss general considerations on the origin of the data, such as the level of knowledge about the author(s) and the data generation process. Usually, researchers are interested that their text corpus comes from the units of analysis they want to make statements about (i.e., social media users) and not about automated bots or other unknown units.

Another example of relevant design decisions relates to the *justification of the level of analysis* (I.6) at which the construct should be measured, such as the sentence, paragraph, document, or corpus level. Here, researchers need to rule out a misalignment between the presumed manifestation of the construct and the level of measurement (as defined by the operationalization), which will likely lead to inconsistent and theory-agnostic measures (Jankowski & Huber, 2022; McKenny et al., 2013).

### *II. Structural Evidence*

For structural evidence, researchers can conduct validation steps to examine and evaluate the properties of the model and its output. Thus, the underlying motivation behind structural evidence is to understand the characteristics of the measurement to eliminate biases and identify limitations. Structural evidence thus enables researchers to gain a deeper understanding of the measurement process. This may require an iterative procedure, where the initial model or features of the textual data are adapted.

Table 2 displays the validation steps that provide structural evidence.

*Table 2: Validation Steps for Structural Evidence (context-independent)*

| ID | Validation Steps | Considerations | Implementation | Applications |
|---|---|---|---|---|
| | | **Model Properties** | | |
| **Model Feature Inspection** | | | | |
| **II.1** | **Inspection of predictive model features** | Have I inspected the predictive features for my model? Have I assured they are | Qualitative evaluation of top-ranked model features | Molnar (2020), |
| | | conceptually aligned with the construct being measured? | using feature-importance methods like e.g., LIME or ICE | Küpfer & Meyer (2023) |
| **Model Output** | | | | |
| Descriptive Output Inspection | | | | |
| **II.2** | **Visual inspection of output** | Have I visualized my output descriptively? Have I identified and visualized outliers and extreme values? | Plotting descriptive statistics; discussing the plausibility of the observed distribution | Goet (2019) |
| **II.3** | **Comparison of aggregated measures across known groups** | Have I aggregated the output scores across known groups (e.g., mean share of sexist sentences across social media user demographics)? | Plotting aggregated measures across groups; discussing the plausibility of the observed distribution | Goet (2019) |
| **II.4** | **Qualitatively assess top documents with the highest overall scores for each output category** | Have I assessed the most outstanding documents for each type of output, such as labels with the highest confidence, or highest and lowest scores on a numerical scale? | Qualitative evaluation to ensure that the top-ranked texts align with the construct | Goet (2019) |
| Error Analysis | | | | |
| **II.5** | **Error analysis using data grouping** | Have I conducted error analysis to compare the performance of my model across known subgroups? | Comparing performance metrics (i.e., F1) across subgroups | Wu et al. (2019) |
| **II.6** | **Error analysis of outstanding or deliberatively chosen observations** | Have I conducted error analysis to qualitatively evaluate the sources and types of errors associated with the measures? | Exploring the underlying causes of misclassifications by qualitatively screening misclassified examples | (Wu et al., 2019) |

*Note: An interactive version of the checklist is available online at https://www.valitex.info*

***Model feature inspection***. The goal of *model feature inspection* (II.1) is to assess whether the characteristics and features of the model are plausible indicators for the construct. For instance, a model aiming to measure populism should prioritize meaningful words like "elite" or "establishment," rather than irrelevant ones unrelated to the construct of interest. Whenever possible, validation can therefore include the inspection of the learned weights of features and distributions (see Linardatos et al., 2020; Molnar, 2020).

***Descriptive output inspection***. Researchers should conduct validation steps to ensure that the measures and their descriptive statistics look plausible. For text analysis, this form of validation steps is often referred to as “face validity” as it solely includes argument-based validation steps that require rather qualitative reasoning and less formal analysis (Goet, 2019; Temporão et al., 2018).[3]

For example, descriptive output inspection can include the *visual inspection of the output* (II.2), in particular outliers and extreme values to get a feeling on the distribution of output scores. Based on this initial examination, researchers can already discuss the plausibility of observed distribution, and identify inconsistencies. Furthermore, researchers can also inspect the *aggregated output across known groups* (II.3) visually. These groups can relate to various variables typically included in the data, such as specific time periods or demographic characteristics (e.g., gender or age groups of social media users). For example, if prior research suggests that men are more likely to exhibit sexist behavior than women, the average proportion of sexist tweets should be higher for men than for women in a sample of social media users. Likewise, it can also be useful to *qualitatively assess the top documents with the highest overall scores for each output category* (II.4), such as for distinct groups or topics, or highest and lowest scores on a numerical scale.

***Error analysis.*** Additionally, researchers should apply error analysis to ensure that systematic biases and errors are considered and evaluated. We here use error analysis as an umbrella term for a set of exploratory analyses that attempt to analyze where mistakes in the text analysis pipeline emerge and how they might affect the outcomes (Wu et al., 2019).

There are broadly two approaches to error analysis. On the one hand, researchers can conduct *error analysis using data grouping* (II.5) to identify especially problematic categories (per-label performance) (Alsallakh et al., 2014). On the other hand, researchers can *conduct qualitative error analysis of outstanding or deliberatively chosen observations* (II.6) to manually identify problematic text characteristics (for an application, see Wadhwa et al., 2018).

[3] In other domains, face validity often refers to the subjective appearance of whether a method appears to be valid (Flake et al., 2017). Due to this ambiguity, we step back from using the term face validity for ValiText, but rather stick to the label "visual inspection of model output"

In Appendix 2, we further outline additional context-specific types of validation evidence for structural evidence, in particular the evaluation of *model metrics* (i.e., reporting of method specific metrics and thresholds) and the implementation of *systemic tests* (i.e., various strategies and experimental approaches to demonstrate that the models can generate meaningful measures for the construct being measured).

### *III. External Evidence*

For external evidence, researchers should conduct validation that test for how the measures corresponds to independent information or criteria. Thus, information outside the scope of the textual data in which the measure was constructed serves as an external benchmark (hence "external" evidence). However, it is important to note that external validation alone is not sufficient because even a model with strong bias (lacking substantive and structural evidence) may exhibit meaningful but misleading associations with external criteria. Hence, focusing excessively on output optimization might result in overlooking errors without a complete understanding of their nature.

Table 3 displays the validation step that provide external evidence.

*Table 3: Validation Steps for External Evidence (context-independent)*

| ID | Validation Steps | Considerations | Implementation | Reference |
|---|---|---|---|---|
| **Measure Interrelation** | | | | |
| | | **Human-annotated test set comparison** | | |
| **III.1** | **Comparison of measures with human-annotated test set ("gold-standard data")** | Have I reached sufficient predictive performance on a test set of held-out human annotations? Did I apply cross-validation to calculate average performance metrics? | Evaluating performance metrics (i.e., F1) for dataset of human annotations | (Samory et al., 2021) |

*Note: An interactive version of the checklist is available online at https://www.valitex.info*

Primarily, researchers should *compare measures with a human-annotated test set* (III.1), often referred to as benchmark or "gold-standard" data (Lowe & Benoit, 2013). Generally, human annotations are considered the most reliable form of validation for text-based research, regardless of the measurement design. In the checklist, we offer additional references to methodological standards and best practices, including coding procedures (using a codebook or pairwise comparison) and reporting classification metrics, such as (average) accuracy, precision, recall and F1-score.

In Appendix 2, we further outline additional context-specific types of validation evidence for external evidence, in particular *surrogate label comparison* (i.e., correspondence to labels that can be linked with the textual data) and *criterion prediction* (i.e., correspondence to external criteria or real-world phenomena).

*IV. Robustness Checks*

Next to the three types of validation evidence outlined above, the ValiText framework also recommends the continuous test of robustness checks to assess the impact of researchers' degree of freedom on the measurement outcome. On a general note, one could see robustness checks as additional means to test whether decisions regarding the measurement design might have a sustainable effect on the measure's outcome.

Table 4 provides an overview over relevant aspects which might be tested.

*Table 4: Robustness Checks (context-independent)*

| ID | Validation Steps | Considerations | Implementation | Reference |
|---|---|---|---|---|
| **Robustness Checks** | | | | |
| **IV.1** | **Rerunning the analysis using alternative text models** | Have I rerun the analysis with alternative text-based methods, such as a baseline model? | Displaying performance metrics (e.g., F1 score on human annotated test set (see III.1)) for alternative measurements | (Samory et al., 2021) |
| **IV.2** | **Rerunning the analysis using different hyperparameter settings** | Have I rerun the analysis with alternative hyperparameter settings? | Displaying performance metrics (e.g., F1 score on human annotated test set (see III.1)) for alternative hyperparameter settings | (Arnold et al., 2023) |
| **IV.3** | **Rerunning the analysis using different cutoff thresholds** | Have I rerun the analysis with alternative cutoff thresholds? | Displaying performance metrics (e.g., F1 score on human annotated test set (see III.1)) for alternative cutoff thresholds | Grimmer et al. (2022) |
| **IV.4** | **Rerunning the analysis using different text cleaning and preprocessing steps** | Have I rerun the analysis using alternative data cleaning and preprocessing settings (e.g., removing certain phrases or features of the data)? | Displaying performance metrics (e.g., F1 score on human annotated test set (see III.1)) for alternative preprocessing steps | Grimmer et al. (2022) |

*Note: An interactive version of the checklist is available online at https://www.valitex.info*

In general, factors to consider for robustness checks relate to rerunning the analysis using *alternative text models* (IV.1), *hyperparameter settings* (IV.2), *cutoff thresholds* (IV.3) and *text cleaning and preprocessing steps* (IV.4) and displaying variance in measurement outcomes and performance metrics.

### *Checklist*

For each of the validation steps presented, we complement our framework with a checklist that is adaptable for different use cases (see Table 5). In particular, we provide individual checklists for applying dictionaries (A), (semi-) supervised classification (B), prompt-base classification using generative LLMs (C) and topic modelling (D).

*Table 5: Use Cases*

| # | Use Case | Training / finetuning data required | Known output categories | Description | Example |
|---|---|---|---|---|---|
| **A** | **Dictionaries** | No | Yes | Assign scores to text units using predefined word lists | A dictionary assigns polarity values ranging from -1 to 1 to each known text unit |
| **B** | **(Semi-) Supervised Classification** | Yes | Yes | Train a model to predict known output categories based on labelled training/fine-tuning data | A pretrained BERT model is fine-tuned on labelled social media posts and predicts “offensive” and “non-offensive” posts |
| **C** | **Prompt-based classification using LLMs** | Yes | Yes | Ask (“prompt”) LLMs to assign labels to texts | A Mistral model is prompted to determine whether a text is “sexist” or “non-sexist” |
| **D** | **Topic Modelling** | No | No | Assign topics without any labeled data | An LDA topic model generates 13 coherent topics |

An important condition for all use cases is that researchers must have full control over their models and be theoretically capable of replicating findings by setting random seeds and having access to model parameters. As a result, we deliberately decided not to include a use case for closed-source blackbox APIs (such as Perspective API, ChatGPT[4] interface, among others) and we generally avoid using them to measure social constructs (Rauchfleisch & Kaiser, 2020; Schoch et al., 2023).

The checklists are provided as fillable templates in Appendix 3 and can also be downloaded from a publicly available shiny app.[5] In Appendix 4, we demonstrate the applicability of the checklists by revisiting and documenting the validation steps from a study by Samory et al. (2019), which measured sexism in social media comments for the use case "B: (Semi-) Supervised Classification". Ultimately, the checklists can be used to it can function as a methodological tool, offering structured guidance for researchers to navigate validation. On the other hand, can also serve as a documentation scheme, streamlining communication between researchers engaged in validation and research consumers.

[4] Lately, providers such as OpenAI allow researchers to set random seeds and system fingerprints when using models via API (https://platform.openai.com/docs/guides/text-generation/reproducible-outputs) to improve reproducibility. However, as the model weights of most models are not public, severe challenges for reproducibility still persist.
[5] Link (anonymized)

## 6. Critical Reflection

We developed the ValiText framework with the intention of tackling the manifold challenges associated with validating text-based measures of social constructs. To advance the overall quality of validation practices beyond the framework proposed, however, we want to emphasize some general conditions for successful validation in greater detail.

**Transparency and Reproducibility**. First and foremost, we want to stress the role of transparency in the validation process. Given that computational text-based measures are susceptible to errors and biases inherent to the models themselves, we argue that researchers should openly acknowledge and embrace the potential limitations and biases of their measurement methods. In practical terms, transparency should entail rigorously documenting the validation process and providing access to the data and code utilized. For example, there is a considerable drive in the NLP literature towards documentation standards, both for the textual data (Bender & Friedman, 2018; Gebru et al., 2021; Heger et al., 2022) and the respective measurement model and its parameters (Derczynski et al., 2023; Dodge et al., 2019; Rogers et al., 2021). Moreover, to address the inherent research freedom associated with text analysis, researchers should anchor their work within robust open science standards, particularly embracing preregistration and registered reports, while also establishing the necessary infrastructure to facilitate replication studies (Schoch et al., 2023).

**Human-in-the-loop**. Second, we are convinced that, at least for now, any computational text-based method should still rely on human semantic understanding of language as an absolute benchmark (van Atteveldt et al., 2021; Weber et al., 2018). Human judgement is crucial because too often, computational methods are prone to rely on spurious relations or noise in the data, thereby lacking a deeper ontological sense of error which should prevent trusting empirical measures blindly (see Jankowski & Huber, 2022). However, human annotations do not automatically lead to accurate labels for (training and) testing models. On the contrary, without adequate guidance and training, human annotation bias might results in low-quality data labels (see Geva et al., 2019; Liu et al., 2021), which might be further aggravated by coders' limited focus, fatigue, and evolving interpretations of the underlying conceptual categories (Neuendorf, 2017). Nevertheless, until today, human understanding remains the most dependable form of annotation that is not easily replaced in the validation process.

**Adaptability and Flexibility.** Although our proposed framework offers a general validation workflow that is broadly applicable, there may be a need to customize validation strategies and have differing viewpoints on what constitutes sufficient validation evidence in specific contexts. For instance, the way meaning is encoded can vary significantly between languages in multilingual settings (for a detailed discussion of multilingual challenges in validation, see Lind et al., 2023). In this case, we would recommend providing additional validation evidence to demonstrate cultural and linguistic sensitivity.

In addition, adaptability also applies to determining appropriate cut-off values and metrics, which unfortunately lack universal interpretability. As a case in point, consider the F1 score, with values above 0.8 generally considered good, and values below 0.5 considered poor. However, the interpretation of the metric depends heavily on the nature of the construct and contextual factors within the measurement design. Thus, researchers must be able to interpret and adapt their specific cut-off values, such as by comparing it to related work or picking the best-performing model out of several competing models.

## 7. Conclusion

We do not claim that our proposed framework provides a "holy grail" that sweeps away all concerns in the process of validating text-based measures. Nevertheless, we emphasize that our framework offers a much-needed structure for discussions circling around validation, benefiting both researchers conducting validation and research consumers seeking solid documentation. Likewise, we hope that our taxonomy of validation steps provides a starting point for a more normative discussion on validation. Because, as evidenced by the decade-long history of validity in the social sciences, methodological discussions concerning validation commonly echo dominant perspectives and best practices within research communities. Therefore, we strongly advocate for community efforts and critical discussion to agree on generally acknowledged standards for the validation of text-based measures of social constructs. Because increased confidence about the validity of text-based measures will not only aid researchers in presenting their research, but also enable various stakeholders – from public institutions to individual researchers– to evaluate and rely on empirical evidence derived from text-based research, thus enhancing the overall credibility and impact of empirical studies.

**Acknowledgments:**

We want to thank Thomas Knopf, Matthias Blümke, Benjamin Guiddenau, Leon Fröhling, Arthur Spirling, Verena Kunz, Fabienne Lind, Daniel Thiele, Carolin Zorell, Matthias Roth, and Martijn Schoonvelde for their insightful feedback.

**Disclosure statement:**

No potential conflict of interest was reported by the author(s).

**Funding:**

This research is partly supported by the Leibniz-Foundation Grant "Digital Dehumanization: Measurements, Exposure and Prevalence"

## 8. Appendix

*Appendix 1: Glossary of Validity Terms*

| Validity Term | (Sub-)type(s) of Evidence | Validation steps |
|---|---|---|
| Face Validity | **Substantive Evidence**<br>• Construct Definition and Operationalization<br>• Design Decisions<br><br>**Structural Evidence**<br>• Descriptive Output Inspection | I.1, I.2, I.4, I.5, I.6, I.7 |
| Content Validity | **Substantive Evidence**<br>• Construct Definition and Operationalization<br>• Design Decisions | I.1, I.2, I.3 |
| Semantic Validity | **Substantive Evidence**<br>• Construct Definition and Operationalization<br>• Design Decisions | I.1, I.2, I.3 |
| Discriminant Validity | **Structural Evidence**<br>• Systematic Testing | II.10 |
| Convergent Validity | **Structural Evidence**<br>• Systematic Testing<br><br>**External Evidence**<br>• Human-Annotated Test Set Comparison<br>• Surrogate Label Comparison | II.11, III.1, III.2 |
| Concurrent Validity | **External Evidence**<br>• Human-Annotated Test Set Comparison<br>• Surrogate Label Comparison | II.11, III.1, III.2 |
| Predictive Validity | **External Evidence**<br>• Criterion Prediction | III.3 |
| Criterion Validity | **External Evidence**<br>• Criterion Prediction | III.3 |
| Hypothesis Validity | **External Evidence**<br>• Criterion Prediction | III.3 |
| Construct Validity | **Substantive Evidence**<br>• Construct Definition and Operationalization<br>• Design Decisions | I.1, I.2, I.4, I.5, I.6, I.7 |

*Appendix 2: Context-dependent Types of Validation Evidence*

Table 6 and 7 present an overview over context-specific validation steps that can further supplement validation efforts. Broadly construed, the *context* can be determined by either the research interest characteristics or the method employed in a particular use case.

### *Context: Data Characteristics*

Table 6 displays the context-specific validation steps for varying *research interest* characteristics.

*Table 6: Validation Steps for research interest characteristics (context-specific)*

| ID | Validation Steps | Context | Considerations | Implementation | Applica-tions |
|---|---|---|---|---|---|
| **Structural Evidence** | | | | | |
| Systematic Testing (context-specific) | | | | | |
| **V.1** | **Counterfactual tests** | **Research Interest**: Demonstrating sensitivity of the measurement | Have I tested that my model is sensitive to meaningful changes in the text data? | Evaluating performance metrics (i.e., F1) for new dataset of counterfactual examples | (Garg et al., 2019) |
| **V.2** | **Adversarial tests** | **Research Interest**: Demonstrating stability of the measurement | Have I tested that my model is resilient to slight perturbations in the text data? | Evaluating performance metrics (i.e., F1) for new dataset of adversarial examples | (Ribeiro et al., 2018) |
| **V.3** | **Discriminant tests** | **Research Interest**: Demonstrating low correspondence to different constructs | Have I tested that my model is able to distinguish between the construct of interest and similar, but unrelated concepts (e.g., and sexist language)? | Inspecting output scores for a sample of "discriminant" examples | Fang et al. (2023) |
| **V.4** | **Out of domain tests** | **Research Interest:** Focus on demonstrating generalizability | Have I tested that my model is able to generalize to out-of-domain examples? | Evaluating performance metrics (i.e., F1) for new dataset of out-of-domain examples | (Sen et al., 2022) |
| **External Evidence** | | | | | |
| Surrogate Label Comparison (context-specific) | | | | | |
| **V.5** | **Comparison of measures with surrogate labels** | **Research Interest**: Demonstrating correspondence to surrogate labels | Have I reached sufficient predictive performance on the surrogate labels? | Evaluating performance metrics (i.e., F1) for the surrogate labels | Grimmer et al. (2022) |
| Criterion Prediction (context-specific) | | | | | |
| **V.6** | **Criterion Prediction** | **Research Interest:** Demonstrating predictive performance to related criteria | Have I been able to accurately predict real-word phenomena? | Evaluating predictive metrics (i.e., regression coefficient) for the criteria | Grimmer et al. (2022) |

## Structural Evidence

**Systematic Testing**. Depending on their research interest, researchers may apply systematic tests that encompass various strategies and experimental approaches, all aimed at demonstrating that the model can generate meaningful measures for the construct being measured.

For example, researchers can conduct *counterfactual tests* (V.1), i.e., strategically synthesized versions of the original texts to put the model to test. For example, counterfactual examples could include changing the polarity of sentences, or adding negations (Ilyas et al., 2019). Thus, increased vulnerability to these changes could highlight important shortcomings and, thus, threats to validity. Likewise, *adversarial tests* (V.2) can be utilized to ensure that the model is resilient to input specifically designed to fool a text model, i.e., by adding slight variations to the original data.[6] In a similar direction, researchers can also draft *discriminant tests* (V.3) to ensure that the model is able to distinguish the construct of interest from similar, but unrelated concepts. For instance, a classifier that is designed to detect benevolent sexism (i.e., subjectively positive view towards men and woman, see (Jha & Mamidi, 2017)) should not necessarily be correlated with hostile sexism (i.e., explicitly negative attitude towards men and woman), as both concepts are assumed to be independent subdimensions of the overall construct of sexism. On a different note, research can also apply *out of domain tests* (V.4) to explore the model's ability to generalize to examples come from a different data distribution that the ones a model has seen before.

[6] https://textattack.readthedocs.io/en/master/

## External Evidence

**Surrogate Labels Comparison**: Often, the data comes with relevant information that can be linked to the construct being measured. Therefore, researchers might further compare their output scores with *surrogate labels* (V.5) that constitute similar, but closely related characteristics of the data, such as the party label as a proxy measure for a politician's speech ideology.

**Criterion Prediction**. In the same vein, researchers might include a validation step to verify that their output scores can forecast *external criteria or real-world phenomena* (V.6). This aims to confirm hypothesized relationships with real-world phenomena, like predicting voting behavior based on textual measures of political ideology (Lauderdale & Herzog, 2016; Rheault & Cochrane, 2020) – much like work performance or school achievement have served as external criteria for tests of cognitive ability or personality in the psychometric tradition.

### *Context: Use-Case Characteristics*

Table 7 displays the context-specific validation steps for varying *use cases* characteristics.

*Table 7: Validation Steps for use case characteristics (context-specific)*

| ID | Validation Steps | Context | Considerations | Implementation | A | B | C | D | Applications |
|---|---|---|---|---|---|---|---|---|---|
| **Substantive Evidence** | | | | | | | | | |
| **Design Decisions** | | | | | | | | | |
| **C.1** | **Including examples into the prompt** | **Method**: Selecting Use Case C: Prompting | Have I supplied clear examples of the construct in the prompt? | Outlining the rationale behind the provided example | - | - | ✓ | - | (Gilardi et al., 2023) |
| **Structural Evidence** | | | | | | | | | |
| **Model Metrics Evaluation** | | | | | | | | | |
| **D.1** | **Evaluation of topic coherence metrics** | **Method**: Selecting Use Case D: Topic Modelling | Have I evaluated topic coherence metrics to review co-occurrence of top words? | Calculating topic coherence metrics (i.e., CV) to select the most coherent topics | - | - | - | ✓ | (Röder et al., 2015) |
| **A.1** | **Evaluation of number of (N-)tokens matched** | **Method**: Selecting Use Case A: Dictionaries | Have I checked the numbers of matches for my dictionary? | Calculating the mean share of matched words; implementing methods to increase matching (e.g., stemming) | ✓ | - | - | - | Goet (2019) |
| **Robustness Checks** | | | | | | | | | |
| **C.2** | Rerunning the analysis using different prompts | **Method**: Selecting Use Case C: Prompting | Have I rerun the analysis with alternative prompt designs? | Displaying performance metrics (e.g., F1 score on human annotated test set (see III.1)) for alternative prompts | - | - | ✓ | - | (Gilardi et al., 2023) |

*Note: A: Dictionaries, B: (Semi-) Supervised Classification, C: Prompt-based classification using LLMs, D: Topic Modelling*

For the use case *A. Dictionaries,* researchers can critically *evaluate the number of words matched by the dictionary* (A.1) as a model metric evaluation. If only a fraction of relevant terms is matched, the researcher might consider implementing methods to increase the number of words matched, such as removing text features or word stemming.

For the use case *C. Prompting*, researchers should provide the *prompts with clear and concrete examples* (C.1) of the construct being measured for substantive evidence. Furthermore, in the robustness checks section, researcher should rerun the analysis using *different prompt templates* and varying prompt designs (C.2).

For the use case *D. Topic Modelling*, researchers can furthermore evaluate *topic coherence metrics* (D.1) that provide guidance on the most coherent number of topics.

*Appendix 3: Illustrative Example: Measuring Sexism in Social Media Comments*

Below, we demonstrate the practical application of the ValiText framework through an illustrative example of measuring sexism in social media comments. We revisit and document validation steps in a study conducted by Samory et al. (2019) that measured sexism in social media comments. Specifically, this study used different types of supervised machine learning models to measure sexism for different datasets. We chose the study by Samory et al. (2019) because of their comprehensive exploration and contemplation of the intricacies tied to measuring sexism. We proceed by scrutinizing the validation steps employed in their study, aligning them with the conceptual framework and the checklist for "B. Classification using Fine-tuned Machine-Learning Model".

***Substantive Evidence***

Starting with *documentation of the conceptual background* (I.1), Samory et al. (2021) provide extensive evidence on their engagement with the relevant literature and other sources of information. For example, the authors discuss existing definitions and attempts to measure sexism using computational methods, conclude that there is definitional unclarity, and reflect on possible biases and spurious artifacts in previous research. Moreover, they evaluate survey measures of sexism (i.e., sets of questions or statements that are used to measure social constructs) from the field of social psychology.

Afterwards, they provide a *justification of the operationalization* (I.2), that is, the link between their construct and the textual data. Because they primarily apply supervised machine-learning methods, the authors' underlying justification lies in the provision of high-quality training data that enables the text model to autonomously learn and adapt relevant patterns in the data. To annotate sexism within their training data, the authors develop a detailed codebook based on four subdimensions of sexism identified in the previous literature (e.g., "behavioral expectations" and "endorsement of inequality"). In addition, they extend the codebook to not only differentiate be-tween sexist content, but also between varying degrees of sexist phrasing. To test their operationalization using a codebook empirically, they rely on *manual pre-coding* (I.3) to label the sexist items from the survey scales into relevant subcategories, finding considerable agreement.

Following this initial stage of construct definition, the authors then discuss their design decisions. They outline their *justification of data collection decisions* (I.4), in particular using different textual datasets. They rely on Twitter data collected through various keywords and strategies and survey scales, while acknowledging the strengths and weaknesses associated with each dataset. In addition, they create a subset of adversarial examples with minimal lexical changes that switches the meaning of sentences from sexist to non-sexist.

Furthermore, for the *justification of method choice* (I.5), the authors rely primarily on a supervised approach. Although implicitly, their argumentation is that only supervised models can replicate human codings that distinguish between different subdimensions of sexism. However, they do not only rely on one specific type of method but rather select a variety of models, such as a Logit model, a CNN and a fine-tuned BERT model with increasing complexity to systematically compare their performance, allowing for a thorough evaluation of measurement performance across different models. Likewise, they also decide to include dictionary baseline models (see the section on robustness checks *rerunning the analysis using alternative text-based methods* (IV.1)).

For justifying the *level of analysis* (I.6), they select the sentence level as the unit of analysis, which is aligned with the literature and the structure of the survey items. For the *justification of preprocessing decisions* (I.7), the authors only provide a detailed description for the Logit model, omitting such details for the other methods.

| ID | Validation Step | Documentation | Considerations | Performance Criteria | Source / References |
|---|---|---|---|---|---|
| **Construct Definition and Operationalization** | | | | | |
| I.1 | Documentation of the conceptual background | • Reference to existing definitions for sexism<br>• Reference to previous attempts to measures sexism (misogyny, benevolent vs. hostile sexism, etc.)<br>• Discussion of implications of definitional unclarity<br>• Reflection on previous models capturing spurious artifacts of the datasets instead of sexist language.<br>• Systematic engagement with survey scales to measures sexist language.<br>• Considerations of sexist phrasing (not only content), i.e., offensive language (no explanation how this decision might relate to literature) | Have I conducted a literature review or consulted with domain experts to gain a sufficient understanding of conceptual background of the construct? | Summarizing existing literature on the conceptual background of the construct | Krippendorf (2018) |
| I.2 | Justification of the operationalization | • Development of a detailed codebook based on four dimensions identified in the psychological literature.<br>• Discussion of coding inconsistencies + Adaption of the coding instructions | Have I sufficiently explained how the construct should manifest itself in the textual data? Have I documented my operationalization in a codebook? | Providing definition and conceptualization of the construct | Krippendorf (2018) |
| I.3 | Manual Precoding | • Test of the codebook using 5 MTURKERS (86% agreement for majority verdict (at least 3 out of 5 agreement) on the survey scales | Have I reached sufficient interrater agreement for a subsample of the textual data? Have I ensured that the construct can be detected in the textual data? Have I outlined my rules of coding uncertainty across coders? | Reaching sufficient interrater agreement (e.g., Krippendorff's alpha α) | Krippendorf (2018), Plank (2022) |
| **Design Decisions** | | | | | |
| I.4 | Justification of data collection decisions | • Combination of different data sources with different characteristics<br>○ Scale items from psychological scales<br>○ Twitter data collected by keywords and human-annotated.<br>○ Twitter data collected by "call me sexist but" phrase (quite experimental approach) | Have I selected a dataset that is representative and relevant to the research question and population of interest? Have I justified the data selection decisions (e.g., using keywords)? Have I | Outlining the rationale behind data selection / collection decisions; Documenting potential | Krippendorf (2018) |
| | | ○ Creation of adversarial examples ("crowd workers to generate adversarial examples, i.e., examples that are a valid input for a machine learning model, strategically synthesized to put the model to test")<br>• Discussion on the features of the annotated datasets | assessed the quality and completeness of the dataset and checked for potential biases or inconsistencies? | limitations and data quality issues | |
| I.5 | Justification of method choice | • Application of different models with increasing complexity (Logit/CNN/Bert) including state of the art methods<br>• Systematic Comparison of these methods | Have I selected the appropriate type of method based on the operationalization of the construct and data characteristics? Have I justified the concrete selection of a particular model? | Outlining the rationale behind method selection; Documenting potential limitations in comparison to alternative methods | Grimmer et al. (2022) |
| I.6 | Justification of the level of analysis | • Not explicitly mentioned, but the focus on the sentence level (based on the survey scales) appears plausible in regard to the literature. | Have I selected the appropriate level of analysis? Have I considered potential problems when aggregating measures from lower to higher levels (e.g., sentence to paragraph level)? | Outline the rationale behind the selected level of analysis (e.g., token, sentence, or paragraph level). | Jankowski & Huber (2022) |
| I.7 | Justification of preprocessing decisions | • Only for the Logit model, a short reference to the adoption of preprocessing decisions similar to Jha and Mamidi is provided. | Have I justified relevant changes to the text prior to the analysis, such as removing certain words or phrases? | Outlining the rationale behind preprocessing decisions | Grimmer et al. (2022) |

### *Structural Evidence*

To demonstrate structural evidence, Samory et al. (2021) proceed with a combination of validation steps to examine and evaluate the properties of the model and its output. To evaluate the model properties, they conduct an *inspection of predictive model features* (II.1). To do so, they evaluate the most predictive words for each sexism category (unigrams) and compare them across their data sets and methods applied. Thus, they observe that some models, which are trained on slightly adapted adversarial examples (see Wallace et al., 2019; Zhang et al., 2019), exhibit more general features, which indicates increased model robustness and, partially, performance.

To evaluate the model output, they furthermore conduct extensive *error analysis using data grouping* (II.5) of misclassified examples to identify systematic errors on their most promising BERT model. They specifically investigate the influence of various factors to assess where the model misclassifies messages using a logistic regression model. Relevant factors which they consider are the type of model used (i.e., whether it was trained on original or adversarial examples), or the origin of the training data. Furthermore, they also evaluate the impact of the initial agreement among the coders on the probability of misclassifying errors.

| ID | Validation Step | Documentation | Considerations | Performance Criteria | Source / References |
|---|---|---|---|---|---|
| Model Feature Inspection | | | | | |
| I.1 | Inspection of predictive model features | • Conducting of feature importance analysis for predictive unigrams (Table 4) | Have I inspected the predictive features for my model? Have I assured they are conceptually aligned with the construct being measured? | Qualitative evaluation of top-ranked model features using feature-importance methods like e.g., LIME or ICE | Molnar (2020), Küpfer & Meyer (2023) |
| Descriptive Output Inspection | | | | | |
| II.2 | Visual inspection of output | • Not provided | Have I visualized my output descriptively? Have I identified and visualized outliers and extreme values? | Plotting descriptive statistics; discussing the plausibility of the observed distribution | Goet (2019) |
| II.3 | Comparison of aggregated measures across known groups | • Not provided | Have I aggregated the output scores across known groups (e.g., mean share of sexist sentences across social media user demographics)? | Plotting aggregated measures across groups; discussing the plausibility of the observed distribution | Goet (2019) |
| II.4 | Qualitatively assess top documents with the highest overall scores for each output category | • Not provided | Have I assessed the most outstanding documents for each type of output, such as labels with the highest confidence, or highest and lowest scores on a numerical scale? | Qualitative evaluation to ensure that the top-ranked texts align with the construct | Goet (2019) |
| Error Analysis | | | | | |
| II.5 | Error analysis using data grouping | • detailed discussion of misclassified examples, identification of systematic errors (e.g., varying performance of baseline model for topicality) | Have I conducted error analysis to compare the performance of my model across known sub-groups? | Comparing performance metrics (i.e., F1) across subgroups | Wu et al. (2019) |
| II.6 | Error analysis of outstanding or deliberatively chosen observations | • Not provided | Have I conducted error analysis to qualitatively evaluate the sources and types of errors associated with the measures? | Exploring the underlying causes of misclassifications by qualitatively screening misclassified examples | (Wu et al., 2019) |
| Systematic Testing (context-specific) | | | | | |
| V.1 | Counterfactual tests | • Conducting counterfactual tests; providing new training samples of counterfactual tests and displaying performance metrics (F1 score). | Have I tested that my model is sensitive to meaningful changes in the text data? | Evaluating performance metrics (i.e., F1) for new dataset of counterfactual examples | (Garg et al., 2019) |
| V.2 | Adversarial tests | • Not provided | Have I tested that my model is resilient to slight perturbations in the text data? | Evaluating performance metrics (i.e., F1) for new dataset of adversarial examples | (Ribeiro et al., 2018) |
| V.3 | Discriminant tests | • Not provided | Have I tested that my model is able to distinguish between the construct of interest and similar, but unrelated concepts (e.g., and sexist language)? | Inspecting output scores for a sample of "discriminant" examples | Fang et al. (2023) |
| V.4 | Out of domain tests | • Not provided | Have I tested that my model is able to generalize to out-of-domain examples? | Evaluating performance metrics (i.e., F1) for new dataset of out-of-domain examples | (Sen et al., 2022) |

### *External Evidence*

Do demonstrate external evidence, Samory et al. (2019) primarily rely on the *comparison of measures with a human-annotated test set* (III.1). To calculate classification performance, they apply k-fold cross-validation and report F1 scores. The evaluation of F1 score is widely regarded as the most viable metric, as alternative metrics such as accuracy (i.e., the overall ratio of positive predictions) can be misleading when dealing with imbalanced data (Spelmen & Porkodi, 2018).

| ID | Validation Step | Documentation | Considerations | Performance Criteria | Source / References |
|---|---|---|---|---|---|
| Construct Definition and Operationalization | | | | | |
| III.1 | Comparison of measures with human-annotated test set ("gold-standard data") | • Systematic comparison with hand-annotated test set, report of F1 scores | Have I reached sufficient predictive performance on a test set of held-out human annotations? Did I apply cross-validation to calculate average performance metrics? | Evaluating performance metrics (i.e., F1) for dataset of human annotations | (Samory et al., 2021) |
| Surrogate Label Comparison (context-specific) | | | | | |
| V.5 | Comparison of measures with surrogate labels | • Not provided | Have I reached sufficient predictive performance on the surrogate labels? | Evaluating performance metrics (i.e., F1) for the surrogate labels | Grimmer et al. (2022) |
| Criterion Prediction (context-specific) | | | | | |
| V.6 | Criterion Prediction | • Not provided | Have I been able to accurately predict real-word phenomena? | Evaluating predictive metrics (i.e., regression coefficient) for the criteria | Grimmer et al. (2022) |

### ***Robustness Checks***

Besides validation steps that provide validation evidence, Samory et al. (2019) conduct a series of robustness checks. For instance, the authors rerun their analysis using alternative text-based methods (IV.1). The inclusion of a simple baseline model (toxicity and gender word dictionary) demonstrates that their supervised models achieve higher overall performance on the hand-annotated test set, providing evidence that their supervised models are more successful in replicating human-annotated measures of sexism.

| ID | Validation Step | Documentation | Considerations | Performance Criteria | Source / References |
|---|---|---|---|---|---|
| Construct Definition and Operationalization | | | | | |
| IV.1 | Rerunning the analysis using alternative text models | • Comparison with several baseline models | Have I rerun the analysis with alternative text-based methods, such as a baseline model? | Displaying performance metrics (e.g., F1 score on human annotated test set (see III.1)) for alternative measurements | (Samory et al., 2021) |
| IV.2 | Rerunning the analysis using different hyperparameter settings | • Not provided | Have I rerun the analysis with alternative hyperparameter settings? | Displaying performance metrics (e.g., F1 score on human annotated test set (see III.1)) for alternative hyperparameter settings | (Arnold et al., 2023) |
| IV.3 | Rerunning the analysis using different cutoff thresholds | • Not provided | Have I rerun the analysis with alternative cutoff thresholds? | Displaying performance metrics (e.g., F1 score on human annotated test set (see III.1)) for alternative cutoff thresholds | Grimmer et al. (2022) |
| IV.4 | Rerunning the analysis using different text cleaning and preprocessing steps | • Not provided | Have I rerun the analysis using alternative data cleaning and preprocessing settings (e.g., removing certain phrases or features of the data)? | Displaying performance metrics (e.g., F1 score on human annotated test set (see III.1)) for alternative preprocessing steps | Grimmer et al. (2022) |

*Appendix 4: Checklist Template*

***Use Case A: Dictionaries***

# Use Case A: Dictionaries

This checklist accompanies the [ValiText] framework for validating text-based measures of social constructs by Birkenmaier et al. (2024). Each row within the table corresponds to one validation step (i.e., specific tests that can be executed to produce validation evidence). As outlined in the corresponding paper, researchers should initially follow the order of the phases, starting with the substantive validation steps and ending with external validation steps while continuously considering robustness checks. However, researchers might adapt this process to their individual use case.

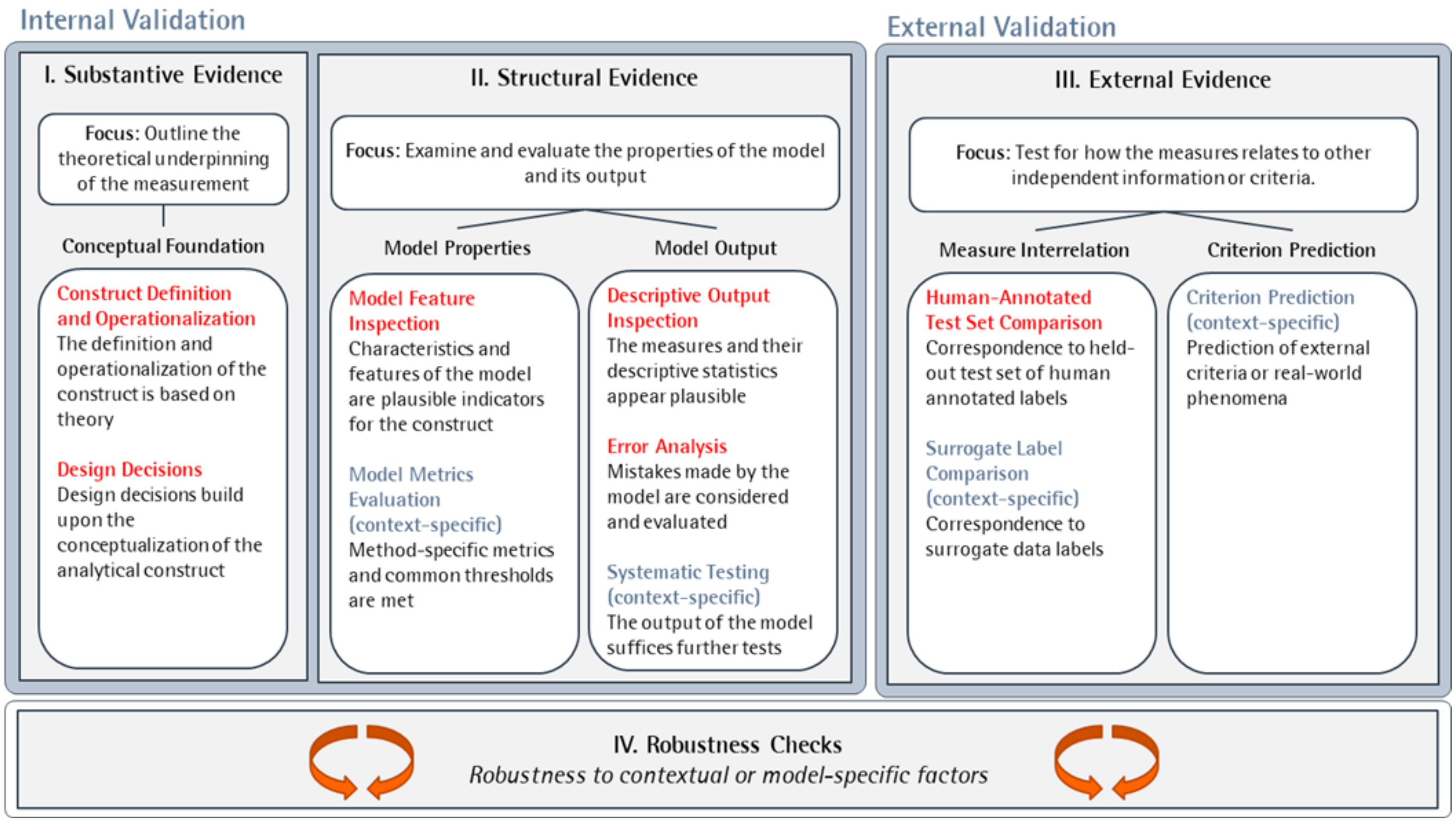

*Figure 3: Framework*

## Substantive Evidence

Before conducting any measurements, researchers need to outline the theoretical under-pinning of the measurement to demonstrate substantive evidence. Validation steps for **substantive evidence** should therefore demonstrate that the measurement is based on a strong conceptual foundation, including the operationalization of the construct and the design decisions around the measurement process.

| ID | Validation Step | Documentation | Considerations | Performance Criteria | Source / References |
|---|---|---|---|---|---|
| **Construct Definition and Operationalization** | | | | | |
| I.1 | Documentation of the conceptual background | • | Have I conducted a literature review or consulted with domain experts to gain a sufficient understanding of conceptual background of the construct? | Summarizing existing literature on the conceptual background of the construct | Krippendorf (2018) |
| I.2 | Justification of the operationalization | • | Have I sufficiently explained how the construct should manifest itself in the textual data? Have I documented my operationalization in a codebook? | Providing definition and conceptualization of the construct | Krippendorf (2018) |
| I.3 | Manual Precoding | • | Have I reached sufficient interrater agreement for a subsample of the textual data? Have I ensured that the construct can be detected in the textual data? Have I outlined my rules of coding uncertainty across coders? | Reaching sufficient interrater agreement (e.g., Krippendorff's alpha α) | Krippendorf (2018), Plank (2022) |
| **Design Decisions** | | | | | |
| I.4 | Justification of data collection decisions | • | Have I selected a dataset that is representative and relevant to the research question and population of interest? Have I justified the data selection decisions (e.g., using keywords)? Have I | Outlining the rationale behind data selection / collection decisions; Documenting potential | Krippendorf (2018) |
| | | | assessed the quality and completeness of the dataset and checked for potential biases or inconsistencies? | limitations and data quality issues | |
| I.5 | Justification of method choice | • | Have I selected the appropriate type of method based on the operationalization of the construct and data characteristics? Have I justified the concrete selection of a particular model? | Outlining the rationale behind method selection; Documenting potential limitations in comparison to alternative methods | Grimmer et al. (2022) |
| I.6 | Justification of the level of analysis | • | Have I selected the appropriate level of analysis? Have I considered potential problems when aggregating measures from lower to higher levels (e.g., sentence to paragraph level)? | Outline the rationale behind the selected level of analysis (e.g., token, sentence, or paragraph level). | Jankowski & Huber (2022) |
| I.7 | Justification of preprocessing decisions | • | Have I justified relevant changes to the text prior to the analysis, such as removing certain words or phrases? | Outlining the rationale behind preprocessing decisions | Grimmer et al. (2022) |

## Structural Evidence

For **structural evidence**, researchers should conduct validation steps to examine and evaluate the properties of the model and its output. Structural evidence enables the researcher to gain a deeper understanding of how the measurement process functions, as well as to identify any biases or errors introduced by the computational workflow.

| ID | Validation Step | Documentation | Considerations | Performance Criteria | Source / References |
|---|---|---|---|---|---|
| Model Feature Inspection | | | | | |
| I.1 | Inspection of predictive model features | • | Have I inspected the predictive features for my model? Have I assured they are conceptually aligned with the construct being measured? | Qualitative evaluation of top-ranked model features using feature-importance methods | Molnar (2020), Küpfer & Meyer (2023) |
| | | | | like e.g., LIME or ICE | |
| Model Feature Inspection | | | | | |
| A.1 | Evaluation of number of (N-)tokens matched | • | Have I checked the numbers of matches for my dictionary? | Calculating the mean share of matched words; implementing methods to increase matching (e.g., stemming) | Goet (2019) |
| Descriptive Output Inspection | | | | | |
| II.2 | Visual inspection of output | • | Have I visualized my output descriptively? Have I identified and visualized outliers and extreme values? | Plotting descriptive statistics; discussing the plausibility of the observed distribution | Goet (2019) |
| II.3 | Comparison of aggregated measures across known groups | • | Have I aggregated the output scores across known groups (e.g., mean share of sexist sentences across social media user demographics)? | Plotting aggregated measures across groups; discussing the plausibility of the observed distribution | Goet (2019) |
| II.4 | Qualitatively assess top documents with the highest overall scores for each output category | • | Have I assessed the most outstanding documents for each type of output, such as labels with the highest confidence, or highest and lowest scores on a numerical scale? | Qualitative evaluation to ensure that the top-ranked texts align with the construct | Goet (2019) |
| Error Analysis | | | | | |
| II.5 | Error analysis using data grouping | • | Have I conducted error analysis to compare the performance of my model across known subgroups? | Comparing performance metrics (i.e., F1) across subgroups | Wu et al. (2019) |
| II.6 | Error analysis of outstanding or deliberatively chosen observations | • | Have I conducted error analysis to qualitatively evaluate the sources and types of errors associated with the measures? | Exploring the underlying causes of misclassifications by qualitatively screening misclassified examples | (Wu et al., 2019) |
| Systematic Testing (context-specific) | | | | | |
| V.1 | Counterfactual tests | • | Have I tested that my model is sensitive to meaningful changes in the text data? | Evaluating performance metrics (i.e., F1) for new dataset of counterfactual examples | (Garg et al., 2019) |
| V.2 | Adversarial tests | • | Have I tested that my model is resilient to slight perturbations in the text data? | Evaluating performance metrics (i.e., F1) for new dataset of adversarial examples | (Ribeiro et al., 2018) |
| V.3 | Discriminant tests | • | Have I tested that my model is able to distinguish between the construct of interest and similar, but unrelated concepts (e.g., and sexist language)? | Inspecting output scores for a sample of "discriminant" examples | Fang et al. (2023) |
| V.4 | Out of domain tests | • | Have I tested that my model is able to generalize to out-of-domain examples? | Evaluating performance metrics (i.e., F1) for new dataset of out-of-domain examples | (Sen et al., 2022) |

## External Evidence

For external evidence, researchers should conduct validation steps that test for how the measures corresponds to independent information or criteria. Thus, information outside the scope of the textual data in which the measure was constructed serves as an external benchmark (hence "external" evidence).

| ID | Validation Step | Documentation | Considerations | Performance Criteria | Source / References |
|---|---|---|---|---|---|
| Construct Definition and Operationalization | | | | | |
| III.1 | Comparison of measures with human-annotated test set ("gold-standard data") | • | Have I reached sufficient predictive performance on a test set of held-out human annotations? Did I apply cross-validation to calculate average performance metrics? | Evaluating performance metrics (i.e., F1) for dataset of human annotations | (Samory et al., 2021) |
| Surrogate Label Comparison (context-specific) | | | | | |
| V.5 | Comparison of measures with surrogate labels | • | Have I reached sufficient predictive performance on the surrogate labels? | Evaluating performance metrics (i.e., F1) for the surrogate labels | Grimmer et al. (2022) |
| Criterion Prediction (context-specific) | | | | | |
| V.6 | Criterion Prediction | • | Have I been able to accurately predict real-word phenomena? | Evaluating predictive metrics (i.e., regression coefficient) for the criteria | Grimmer et al. (2022) |

## Robustness Checks

Next to the three types of validation evidence outlined above, the ValiText framework also recommends the continuous test of robustness checks to assess the impact of researchers' degree of freedom on the measurement outcome. On a general note, one could see robustness checks as additional means to test whether decisions regarding the measurement design might have a sustainable effect on the measure's outcome.

| ID | Validation Step | Documentation | Considerations | Performance Criteria | Source / References |
|---|---|---|---|---|---|
| Construct Definition and Operationalization | | | | | |
| IV.1 | Rerunning the analysis using alternative text models | • | Have I rerun the analysis with alternative text-based methods, such as a baseline model? | Displaying performance metrics (e.g., F1 score on human annotated test set (see III.1)) for alternative measurements | (Samory et al., 2021) |
| IV.2 | Rerunning the analysis using different hyperparameter settings | • | Have I rerun the analysis with alternative hyperparameter settings? | Displaying performance metrics (e.g., F1 score on human annotated test set (see III.1)) for alternative hyperparameter settings | (Arnold et al., 2023) |
| IV.3 | Rerunning the analysis using different cutoff thresholds | • | Have I rerun the analysis with alternative cutoff thresholds? | Displaying performance metrics (e.g., F1 score on human annotated test set (see III.1)) for alternative cutoff thresholds | Grimmer et al. (2022) |
| IV.4 | Rerunning the analysis using different text cleaning and preprocessing steps | • | Have I rerun the analysis using alternative data cleaning and preprocessing settings (e.g., removing certain phrases or features of the data)? | Displaying performance metrics (e.g., F1 score on human annotated test set (see III.1)) for alternative preprocessing steps | Grimmer et al. (2022) |

# Use Case B: (Semi-) Supervised Classification

This checklist accompanies the [ValiText] framework for validating text-based measures of social constructs by Birkenmaier et al. (2024). Each row within the table corresponds to one validation step (i.e., specific tests that can be executed to produce validation evidence). As outlined in the corresponding paper, researchers should initially follow the order of the phases, starting with the substantive validation steps and ending with external validation steps while continuously considering robustness checks. However, researchers might adapt this process to their individual use case.

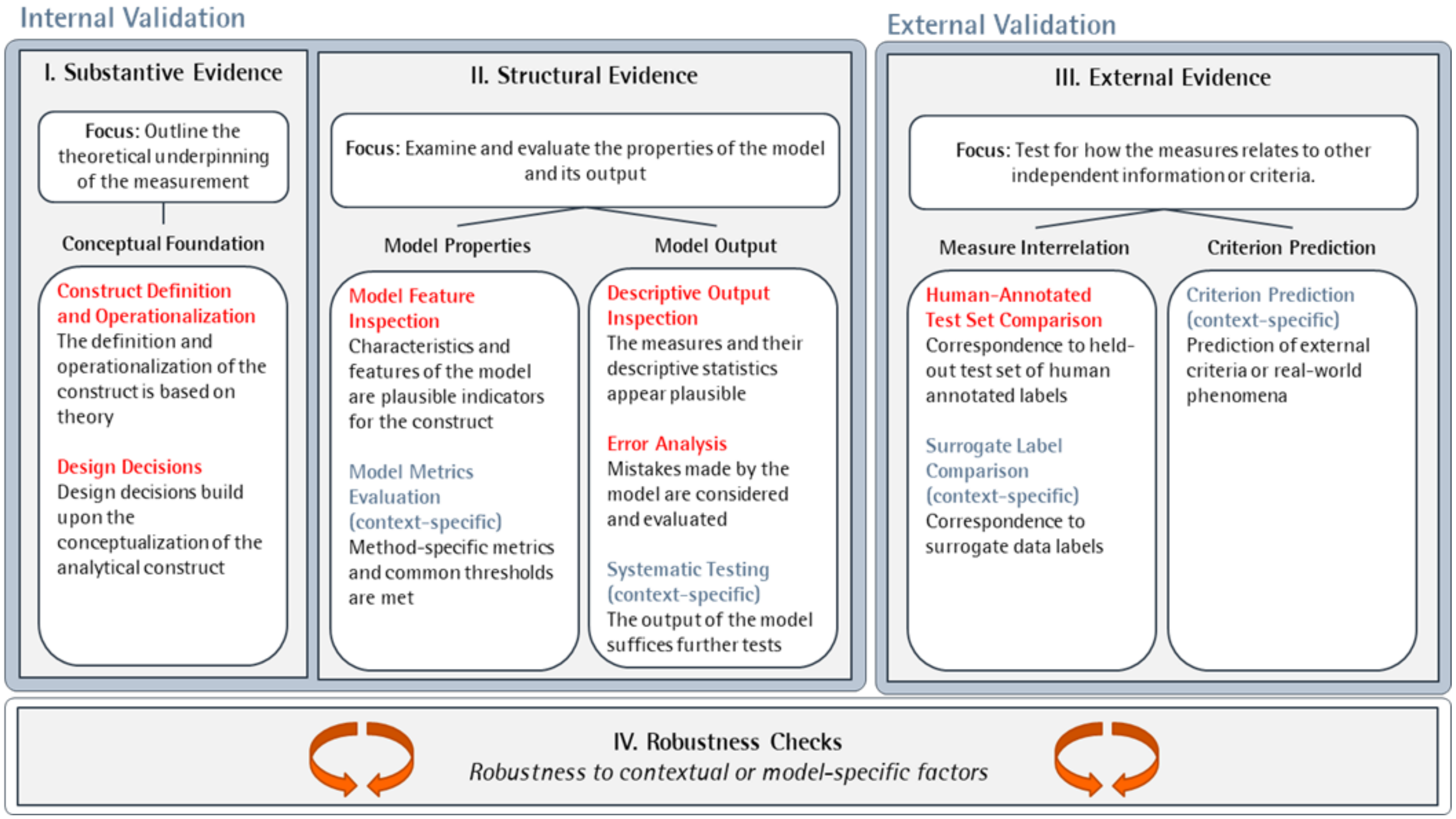

*Figure 4: Framework*

## Substantive Evidence

Before conducting any measurements, researchers need to outline the theoretical under-pinning of the measurement to demonstrate substantive evidence. Validation steps for **substantive evidence** should therefore demonstrate that the measurement is based on a strong conceptual foundation, including the operationalization of the construct and the design decisions around the measurement process.

| ID | Validation Step | Documentation | Considerations | Performance Criteria | Source / References |
|---|---|---|---|---|---|
| **Construct Definition and Operationalization** | | | | | |
| I.1 | Documentation of the conceptual background | • | Have I conducted a literature review or consulted with domain experts to gain a sufficient understanding of conceptual background of the construct? | Summarizing existing literature on the conceptual background of the construct | Krippendorf (2018) |
| I.2 | Justification of the operationalization | • | Have I sufficiently explained how the construct should manifest itself in the textual data? Have I documented my operationalization in a codebook? | Providing definition and conceptualization of the construct | Krippendorf (2018) |
| I.3 | Manual Precoding | • | Have I reached sufficient interrater agreement for a subsample of the textual data? Have I ensured that the construct can be detected in the textual data? Have I outlined my rules of coding uncertainty across coders? | Reaching sufficient interrater agreement (e.g., Krippendorff's alpha $\alpha$) | Krippendorf (2018), Plank (2022) |
| **Design Decisions** | | | | | |
| I.4 | Justification of data collection decisions | • | Have I selected a dataset that is representative and relevant to the research question and population of interest? Have I justified the data selection decisions | Outlining the rationale behind data selection / collection decisions; | Krippendorf (2018) |
| | | | (e.g., using keywords)? Have I assessed the quality and completeness of the dataset and checked for potential biases or inconsistencies? | Documenting potential limitations and data quality issues | |
| I.5 | Justification of method choice | • | Have I selected the appropriate type of method based on the operationalization of the construct and data characteristics? Have I justified the concrete selection of a particular model? | Outlining the rationale behind method selection; Documenting potential limitations in comparison to alternative methods | Grimmer et al. (2022) |
| I.6 | Justification of the level of analysis | • | Have I selected the appropriate level of analysis? Have I considered potential problems when aggregating measures from lower to higher levels (e.g., sentence to paragraph level)? | Outline the rationale behind the selected level of analysis (e.g., token, sentence, or paragraph level). | Jankowski & Huber (2022) |
| I.7 | Justification of preprocessing decisions | • | Have I justified relevant changes to the text prior to the analysis, such as removing certain words or phrases? | Outlining the rationale behind preprocessing decisions | Grimmer et al. (2022) |

## Structural Evidence

For **structural evidence**, researchers should conduct validation steps to examine and evaluate the properties of the model and its output. Structural evidence enables the researcher to gain a deeper understanding of how the measurement process functions, as well as to identify any biases or errors introduced by the computational workflow.

| ID | Validation Step | Documentation | Considerations | Performance Criteria | Source / References |
|---|---|---|---|---|---|
| Model Feature Inspection | | | | | |
| I.1 | Inspection of predictive model features | • | Have I inspected the predictive features for my model? Have I assured they are conceptually aligned with the construct being measured? | Qualitative evaluation of top-ranked model features using feature-importance methods like e.g., LIME or ICE | Molnar (2020), Küpfer & Meyer (2023) |
| Descriptive Output Inspection | | | | | |
| II.2 | Visual inspection of output | • | Have I visualized my output descriptively? Have I identified and visualized outliers and extreme values? | Plotting descriptive statistics; discussing the plausibility of the observed distribution | Goet (2019) |
| II.3 | Comparison of aggregated measures across known groups | • | Have I aggregated the output scores across known groups (e.g., mean share of sexist sentences across social media user demographics)? | Plotting aggregated measures across groups; discussing the plausibility of the observed distribution | Goet (2019) |
| II.4 | Qualitatively assess top documents with the highest overall scores for each output category | • | Have I assessed the most outstanding documents for each type of output, such as labels with the highest confidence, or highest and lowest scores on a numerical scale? | Qualitative evaluation to ensure that the top-ranked texts align with the construct | Goet (2019) |
| Error Analysis | | | | | |
| II.5 | Error analysis using data grouping | • | Have I conducted error analysis to compare the performance of my model across known subgroups? | Comparing performance metrics (i.e., F1) across subgroups | Wu et al. (2019) |
| II.6 | Error analysis of outstanding or deliberatively chosen observations | • | Have I conducted error analysis to qualitatively evaluate the sources and types of errors associated with the measures? | Exploring the underlying causes of misclassifications by qualitatively screening misclassified examples | (Wu et al., 2019) |
| Systematic Testing (context-specific) | | | | | |
| V.1 | Counterfactual tests | • | Have I tested that my model is sensitive to meaningful changes in the text data? | Evaluating performance metrics (i.e., F1) for new dataset of counterfactual examples | (Garg et al., 2019) |
| V.2 | Adversarial tests | • | Have I tested that my model is resilient to slight perturbations in the text data? | Evaluating performance metrics (i.e., F1) for new dataset of adversarial examples | (Ribeiro et al., 2018) |
| V.3 | Discriminant tests | • | Have I tested that my model is able to distinguish between the construct of interest and similar, but unrelated concepts (e.g., and sexist language)? | Inspecting output scores for a sample of "discriminant" examples | Fang et al. (2023) |
| V.4 | Out of domain tests | • | Have I tested that my model is able to generalize to out-of-domain examples? | Evaluating performance metrics (i.e., F1) for new dataset of out-of-domain examples | (Sen et al., 2022) |

## External Evidence

For external evidence, researchers should conduct validation steps that test for how the measures corresponds to independent information or criteria. Thus, information outside the scope of the textual data in which the measure was constructed serves as an external benchmark (hence "external" evidence).

| ID | Validation Step | Documentation | Considerations | Performance Criteria | Source / References |
|---|---|---|---|---|---|
| Construct Definition and Operationalization | | | | | |
| III.1 | Comparison of measures with human-annotated test set ("gold-standard data") | • | Have I reached sufficient predictive performance on a test set of held-out human annotations? Did I apply cross-validation to calculate average performance metrics? | Evaluating performance metrics (i.e., F1) for dataset of human annotations | (Samory et al., 2021) |
| Surrogate Label Comparison (context-specific) | | | | | |
| V.5 | Comparison of measures with surrogate labels | • | Have I reached sufficient predictive performance on the surrogate labels? | Evaluating performance metrics (i.e., F1) for the surrogate labels | Grimmer et al. (2022) |
| Criterion Prediction (context-specific) | | | | | |
| V.6 | Criterion Prediction | • | Have I been able to accurately predict real-word phenomena? | Evaluating predictive metrics (i.e., regression coefficient) for the criteria | Grimmer et al. (2022) |

## **Robustness** Checks

Next to the three types of validation evidence outlined above, the ValiText framework also recommends the continuous test of robustness checks to assess the impact of researchers' degree of freedom on the measurement outcome. On a general note, one could see robustness checks as additional means to test whether decisions regarding the measurement design might have a sustainable effect on the measure's outcome.

| ID | Validation Step | Documentation | Considerations | Performance Criteria | Source / References |
|---|---|---|---|---|---|
| Construct Definition and Operationalization | | | | | |
| IV.1 | Rerunning the analysis using alternative text models | • | Have I rerun the analysis with alternative text-based methods, such as a baseline model? | Displaying performance metrics (e.g., F1 score on human annotated test set (see III.1)) for alternative measurements | (Samory et al., 2021) |
| IV.2 | Rerunning the analysis using different hyperparameter settings | • | Have I rerun the analysis with alternative hyperparameter settings? | Displaying performance metrics (e.g., F1 score on human annotated test set (see III.1)) for alternative hyperparameter settings | (Arnold et al., 2023) |
| IV.3 | Rerunning the analysis using different cutoff thresholds | • | Have I rerun the analysis with alternative cutoff thresholds? | Displaying performance metrics (e.g., F1 score on human annotated test set (see III.1)) for alternative cutoff thresholds | Grimmer et al. (2022) |
| IV.4 | Rerunning the analysis using different text cleaning and preprocessing steps | • | Have I rerun the analysis using alternative data cleaning and preprocessing settings (e.g., removing certain phrases or features of the data)? | Displaying performance metrics (e.g., F1 score on human annotated test set (see III.1)) for alternative preprocessing steps | Grimmer et al. (2022) |

# Use Case C: Prompt-based classification using LLMs

This checklist accompanies the ValiText framework for validating text-based measures of social constructs by Birkenmaier et al. (2024). Each row within the table corresponds to one validation step (i.e., specific tests that can be executed to produce validation evidence). As outlined in the corresponding paper, researchers should initially follow the order of the phases, starting with the substantive validation steps and ending with external validation steps while continuously considering robustness checks. However, researchers might adapt this process to their individual use case.

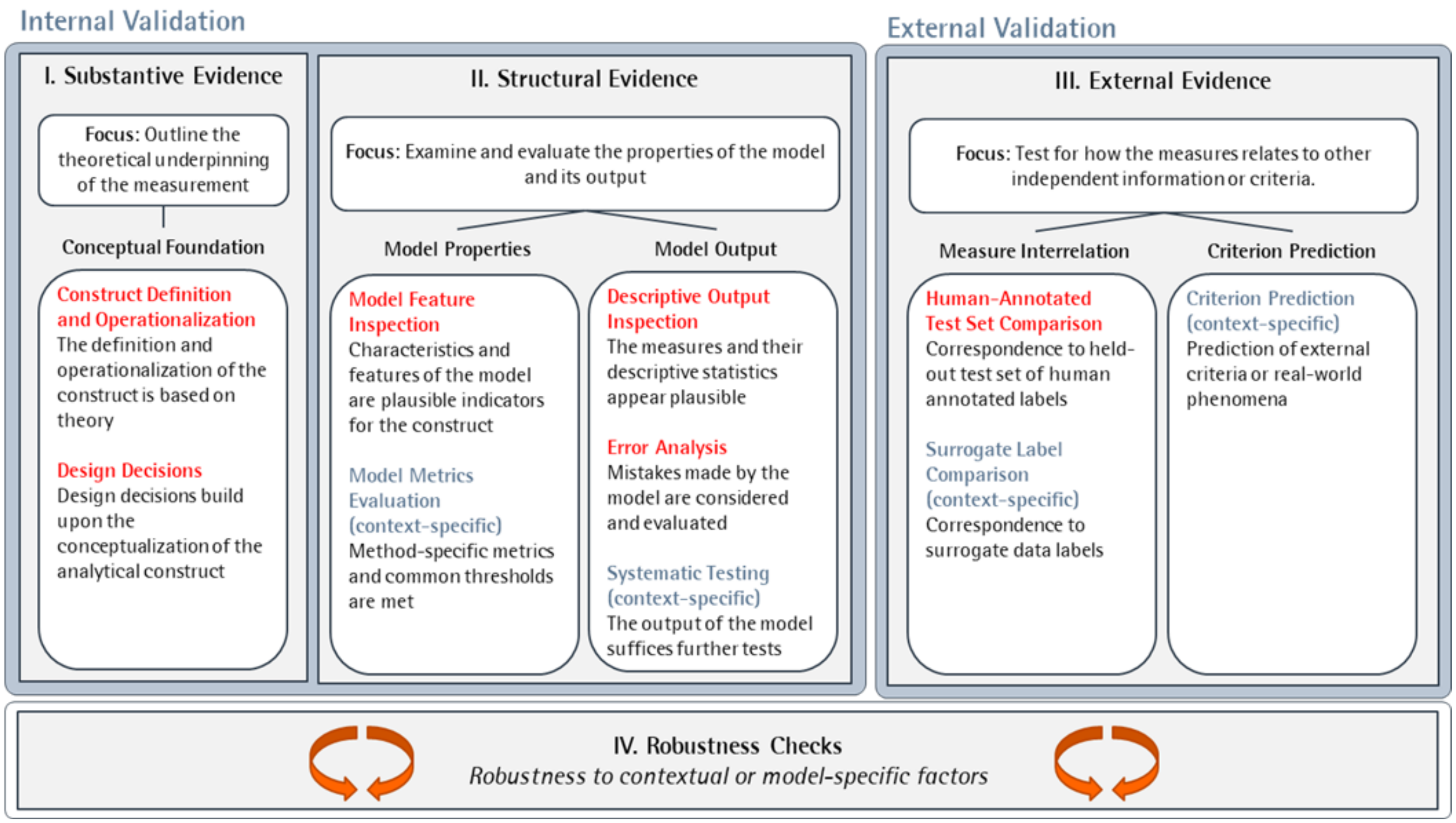

*Figure 5: Framework*

## Substantive Evidence

Before conducting any measurements, researchers need to outline the theoretical under-pinning of the measurement to demonstrate substantive evidence. Validation steps for **substantive evidence** should therefore demonstrate that the measurement is based on a strong conceptual foundation, including the operationalization of the construct and the design decisions around the measurement process.

| ID | Validation Step | Documentation | Considerations | Performance Criteria | Source / References |
|---|---|---|---|---|---|
| **Construct Definition and Operationalization** | | | | | |
| I.1 | Documentation of the conceptual background | • | Have I conducted a literature review or consulted with domain experts to gain a sufficient understanding of conceptual background of the construct? | Summarizing existing literature on the conceptual background of the construct | Krippendorf (2018) |
| I.2 | Justification of the operationalization | • | Have I sufficiently explained how the construct should manifest itself in the textual data? Have I documented my operationalization in a codebook? | Providing definition and conceptualization of the construct | Krippendorf (2018) |
| I.3 | Manual Precoding | • | Have I reached sufficient interrater agreement for a subsample of the textual data? Have I ensured that the construct can be detected in the textual data? Have I outlined my rules of coding uncertainty across coders? | Reaching sufficient interrater agreement (e.g., Krippendorff's alpha α) | Krippendorf (2018), Plank (2022) |
| **Design Decisions** | | | | | |
| C.1 | Including examples into the prompt | • | Have I supplied clear examples of the construct in the prompt? | Outlining the rationale behind the provided example | (Gilardi et al., 2023) |
| I.4 | Justification of data collection decisions | • | Have I selected a dataset that is representative and relevant to the research question and population of interest? Have I justified the data selection decisions (e.g., using keywords)? Have I | Outlining the rationale behind data selection / collection decisions; Documenting potential | Krippendorf (2018) |
| | | | assessed the quality and completeness of the dataset and checked for potential biases or inconsistencies? | limitations and data quality issues | |
| I.5 | Justification of method choice | • | Have I selected the appropriate type of method based on the operationalization of the construct and data characteristics? Have I justified the concrete selection of a particular model? | Outlining the rationale behind method selection; Documenting potential limitations in comparison to alternative methods | Grimmer et al. (2022) |
| I.6 | Justification of the level of analysis | • | Have I selected the appropriate level of analysis? Have I considered potential problems when aggregating measures from lower to higher levels (e.g., sentence to paragraph level)? | Outline the rationale behind the selected level of analysis (e.g., token, sentence, or paragraph level). | Jankowski & Huber (2022) |
| I.7 | Justification of preprocessing decisions | • | Have I justified relevant changes to the text prior to the analysis, such as removing certain words or phrases? | Outlining the rationale behind preprocessing decisions | Grimmer et al. (2022) |

## Structural Evidence

For **structural evidence**, researchers should conduct validation steps to examine and evaluate the properties of the model and its output. Structural evidence enables the researcher to gain a deeper understanding of how the measurement process functions, as well as to identify any biases or errors introduced by the computational workflow.

| ID | Validation Step | Documentation | Considerations | Performance Criteria | Source / References |
|---|---|---|---|---|---|
| Model Feature Inspection | | | | | |
| I.1 | Inspection of predictive model features | • | Have I inspected the predictive features for my model? Have I assured they are conceptually aligned with the construct being measured? | Qualitative evaluation of top-ranked model features using feature-importance methods like e.g., LIME or ICE | Molnar (2020), Küpfer & Meyer (2023) |
| Descriptive Output Inspection | | | | | |
| II.2 | Visual inspection of output | • | Have I visualized my output descriptively? Have I identified and visualized outliers and extreme values? | Plotting descriptive statistics; discussing the plausibility of the observed distribution | Goet (2019) |
| II.3 | Comparison of aggregated measures across known groups | • | Have I aggregated the output scores across known groups (e.g., mean share of sexist sentences across social media user demographics)? | Plotting aggregated measures across groups; discussing the plausibility of the observed distribution | Goet (2019) |
| II.4 | Qualitatively assess top documents with the highest overall scores for each output category | • | Have I assessed the most outstanding documents for each type of output, such as labels with the highest confidence, or highest and lowest scores on a numerical scale? | Qualitative evaluation to ensure that the top-ranked texts align with the construct | Goet (2019) |
| Error Analysis | | | | | |
| II.5 | Error analysis using data grouping | • | Have I conducted error analysis to compare the performance of my model across known subgroups? | Comparing performance metrics (i.e., F1) across subgroups | Wu et al. (2019) |
| II.6 | Error analysis of outstanding or deliberatively chosen observations | • | Have I conducted error analysis to qualitatively evaluate the sources and types of errors associated with the measures? | Exploring the underlying causes of misclassifications by qualitatively screening misclassified examples | (Wu et al., 2019) |
| Systematic Testing (context-specific) | | | | | |
| V.1 | Counterfactual tests | • | Have I tested that my model is sensitive to meaningful changes in the text data? | Evaluating performance metrics (i.e., F1) for new dataset of counterfactual examples | (Garg et al., 2019) |
| V.2 | Adversarial tests | • | Have I tested that my model is resilient to slight perturbations in the text data? | Evaluating performance metrics (i.e., F1) for new dataset of adversarial examples | (Ribeiro et al., 2018) |
| V.3 | Discriminant tests | • | Have I tested that my model is able to distinguish between the construct of interest and similar, but unrelated concepts (e.g., and sexist language)? | Inspecting output scores for a sample of "discriminant" examples | Fang et al. (2023) |
| V.4 | Out of domain tests | • | Have I tested that my model is able to generalize to out-of-domain examples? | Evaluating performance metrics (i.e., F1) for new dataset of out-of-domain examples | (Sen et al., 2022) |

## External Evidence

For external evidence, researchers should conduct validation steps that test for how the measures corresponds to independent information or criteria. Thus, information outside the scope of the textual data in which the measure was constructed serves as an external benchmark (hence "external" evidence).

| ID | Validation Step | Documentation | Considerations | Performance Criteria | Source / References |
|---|---|---|---|---|---|
| Construct Definition and Operationalization | | | | | |
| III.1 | Comparison of measures with human-annotated test set ("gold-standard data") | • | Have I reached sufficient predictive performance on a test set of held-out human annotations? Did I apply cross-validation to calculate average performance metrics? | Evaluating performance metrics (i.e., F1) for dataset of human annotations | (Samory et al., 2021) |
| Surrogate Label Comparison (context-specific) | | | | | |
| V.5 | Comparison of measures with surrogate labels | • | Have I reached sufficient predictive performance on the surrogate labels? | Evaluating performance metrics (i.e., F1) for the surrogate labels | Grimmer et al. (2022) |
| Criterion Prediction (context-specific) | | | | | |
| V.6 | Criterion Prediction | • | Have I been able to accurately predict real-word phenomena? | Evaluating predictive metrics (i.e., regression coefficient) for the criteria | Grimmer et al. (2022) |

# Robustness Checks

Next to the three types of validation evidence outlined above, the ValiText framework also recommends the continuous test of robustness checks to assess the impact of researchers' degree of freedom on the measurement outcome. On a general note, one could see robustness checks as additional means to test whether decisions regarding the measurement design might have a sustainable effect on the measure's outcome.

| ID | Validation Step | Documentation | Considerations | Performance Criteria | Source / References |
|---|---|---|---|---|---|
| Construct Definition and Operationalization | | | | | |
| IV.1 | Rerunning the analysis using alternative text models | • | Have I rerun the analysis with alternative text-based methods, such as a baseline model? | Displaying performance metrics (e.g., F1 score on human annotated test set (see III.1)) for alternative measurements | (Samory et al., 2021) |
| IV.2 | Rerunning the analysis using different hyperparameter settings | • | Have I rerun the analysis with alternative hyperparameter settings? | Displaying performance metrics (e.g., F1 score on human annotated test set (see III.1)) for alternative hyperparameter settings | (Arnold et al., 2023) |
| IV.3 | Rerunning the analysis using different cutoff thresholds | • | Have I rerun the analysis with alternative cutoff thresholds? | Displaying performance metrics (e.g., F1 score on human annotated test set (see III.1)) for alternative cutoff thresholds | Grimmer et al. (2022) |
| IV.4 | Rerunning the analysis using different text cleaning and preprocessing steps | • | Have I rerun the analysis using alternative data cleaning and preprocessing settings (e.g., removing certain phrases or features of the data)? | Displaying performance metrics (e.g., F1 score on human annotated test set (see III.1)) for alternative preprocessing steps | Grimmer et al. (2022) |
| C.2 | Rerunning the analysis using different prompts | • | Have I rerun the analysis with alternative prompt designs? | Displaying performance metrics (e.g., F1 score on human annotated test set (see III.1)) for alternative prompts | (Gilardi et al., 2023) |

# Use Case D: Topic Modelling

This checklist accompanies the ValiText framework for validating text-based measures of social constructs by Birkenmaier et al. (2024). Each row within the table corresponds to one validation step (i.e., specific tests that can be executed to produce validation evidence). As outlined in the corresponding paper, researchers should initially follow the order of the phases, starting with the substantive validation steps and ending with external validation steps while continuously considering robustness checks. However, researchers might adapt this process to their individual use case.

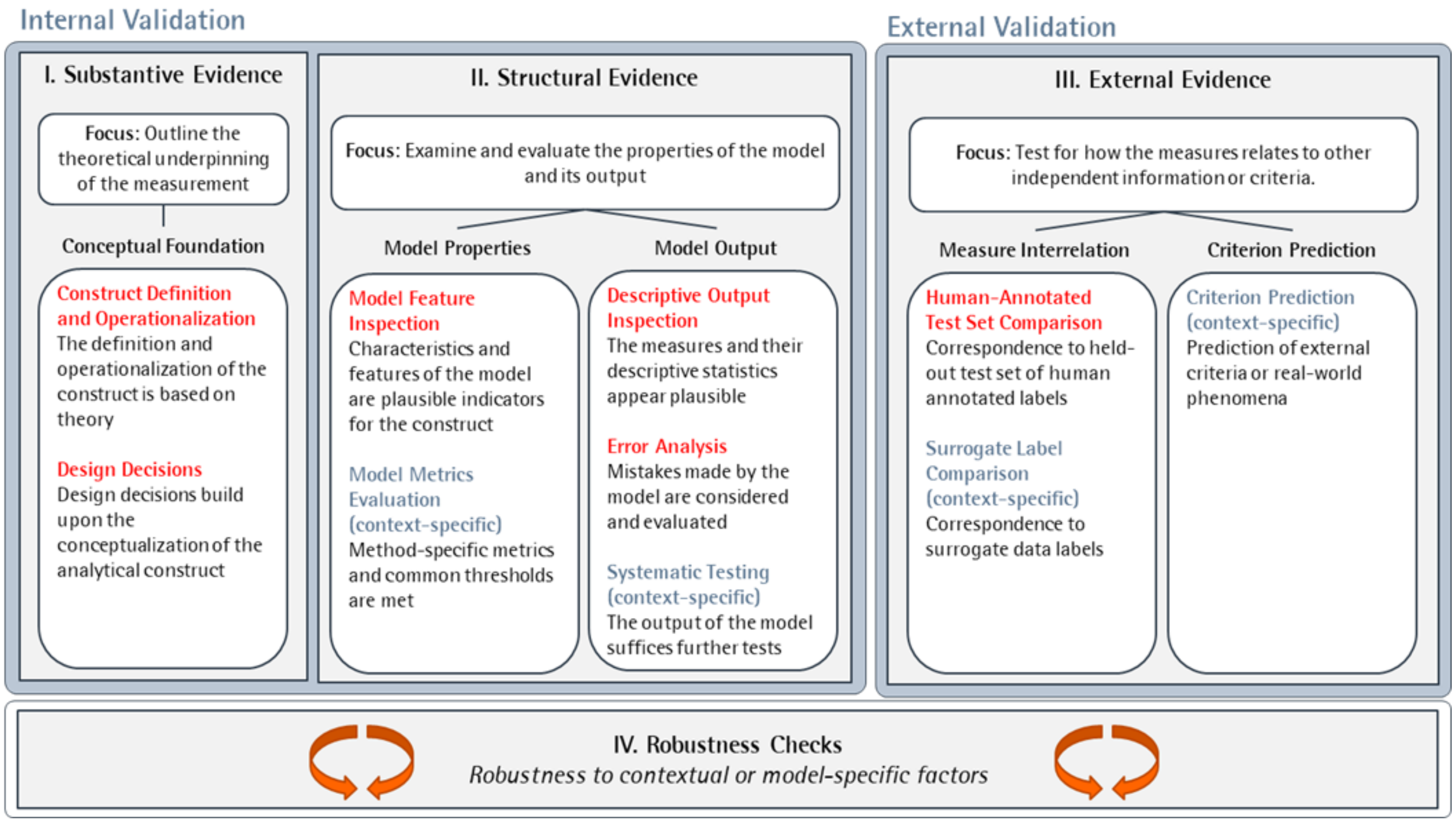

*Figure 6: Framework*

## Substantive Evidence

Before conducting any measurements, researchers need to outline the theoretical under-pinning of the measurement to demonstrate substantive evidence. Validation steps for **substantive evidence** should therefore demonstrate that the measurement is based on a strong conceptual foundation, including the operationalization of the construct and the design decisions around the measurement process.

| ID | Validation Step | Documentation | Considerations | Performance Criteria | Source / References |
|---|---|---|---|---|---|
| **Construct Definition and Operationalization** | | | | | |
| I.1 | Documentation of the conceptual background | • | Have I conducted a literature review or consulted with domain experts to gain a sufficient understanding of conceptual background of the construct? | Summarizing existing literature on the conceptual background of the construct | Krippendorf (2018) |
| I.2 | Justification of the operationalization | • | Have I sufficiently explained how the construct should manifest itself in the textual data? Have I documented my operationalization in a codebook? | Providing definition and conceptualization of the construct | Krippendorf (2018) |
| I.3 | Manual Precoding | • | Have I reached sufficient interrater agreement for a subsample of the textual data? Have I ensured that the construct can be detected in the textual data? Have I outlined my rules of coding uncertainty across coders? | Reaching sufficient interrater agreement (e.g., Krippendorff's alpha α) | Krippendorf (2018), Plank (2022) |
| **Design Decisions** | | | | | |
| I.4 | Justification of data collection decisions | • | Have I selected a dataset that is representative and relevant to the research question and population of interest? Have I justified the data selection decisions (e.g., using keywords)? Have I assessed the quality and completeness of the dataset and | Outlining the rationale behind data selection / collection decisions; Documenting potential limitations and data quality issues | Krippendorf (2018) |
| | | | checked for potential biases or inconsistencies? | | |
| I.5 | Justification of method choice | • | Have I selected the appropriate type of method based on the operationalization of the construct and data characteristics? Have I justified the concrete selection of a particular model? | Outlining the rationale behind method selection; Documenting potential limitations in comparison to alternative methods | Grimmer et al. (2022) |
| I.6 | Justification of the level of analysis | • | Have I selected the appropriate level of analysis? Have I considered potential problems when aggregating measures from lower to higher levels (e.g., sentence to paragraph level)? | Outline the rationale behind the selected level of analysis (e.g., token, sentence, or paragraph level). | Jankowski & Huber (2022) |
| I.7 | Justification of preprocessing decisions | • | Have I justified relevant changes to the text prior to the analysis, such as removing certain words or phrases? | Outlining the rationale behind preprocessing decisions | Grimmer et al. (2022) |

## Structural Evidence

For **structural evidence**, researchers should conduct validation steps to examine and evaluate the properties of the model and its output. Structural evidence enables the researcher to gain a deeper understanding of how the measurement process functions, as well as to identify any biases or errors introduced by the computational workflow.

| ID | Validation Step | Documentation | Considerations | Performance Criteria | Source / References |
|---|---|---|---|---|---|
| Model Feature Inspection | | | | | |
| I.1 | Inspection of predictive model features | • | Have I inspected the predictive features for my model? Have I assured they are conceptually aligned with the construct being measured? | Qualitative evaluation of top-ranked model features using feature-importance methods like e.g., LIME or ICE | Molnar (2020), Küpfer & Meyer (2023) |
| Model Metrics Evaluation (context-specific) | | | | | |
| D.1 | Evaluation of topic coherence metrics | • | Have I evaluated topic coherence metrics to review co-occurrence of top words? | Calculating topic coherence metrics (i.e., CV) to select the most coherent topics | (Röder et al., 2015) |
| Descriptive Output Inspection | | | | | |
| II.2 | Visual inspection of output | • | Have I visualized my output descriptively? Have I identified and visualized outliers and extreme values? | Plotting descriptive statistics; discussing the plausibility of the observed distribution | Goet (2019) |
| II.3 | Comparison of aggregated measures across known groups | • | Have I aggregated the output scores across known groups (e.g., mean share of sexist sentences across social media user demographics)? | Plotting aggregated measures across groups; discussing the plausibility of the observed distribution | Goet (2019) |
| II.4 | Qualitatively assess top documents with the highest overall scores for each output category | • | Have I assessed the most outstanding documents for each type of output, such as labels with the highest confidence, or highest and lowest scores on a numerical scale? | Qualitative evaluation to ensure that the top-ranked texts align with the construct | Goet (2019) |
| Error Analysis | | | | | |
| II.5 | Error analysis using data grouping | • | Have I conducted error analysis to compare the performance of my model across known subgroups? | Comparing performance metrics (i.e., F1) across subgroups | Wu et al. (2019) |
| II.6 | Error analysis of outstanding or deliberatively chosen observations | • | Have I conducted error analysis to qualitatively evaluate the sources and types of errors associated with the measures? | Exploring the underlying causes of misclassifications by qualitatively screening misclassified examples | (Wu et al., 2019) |
| Systematic Testing (context-specific) | | | | | |
| V.1 | Counterfactual tests | • | Have I tested that my model is sensitive to meaningful changes in the text data? | Evaluating performance metrics (i.e., F1) for new dataset of counterfactual examples | (Garg et al., 2019) |
| V.2 | Adversarial tests | • | Have I tested that my model is resilient to slight perturbations in the text data? | Evaluating performance metrics (i.e., F1) for new dataset of adversarial examples | (Ribeiro et al., 2018) |
| V.3 | Discriminant tests | • | Have I tested that my model is able to distinguish between the construct of interest and similar, but unrelated concepts (e.g., and sexist language)? | Inspecting output scores for a sample of "discriminant" examples | Fang et al. (2023) |
| V.4 | Out of domain tests | • | Have I tested that my model is able to generalize to out-of-domain examples? | Evaluating performance metrics (i.e., F1) for new dataset of out-of-domain examples | (Sen et al., 2022) |

## External Evidence

For external evidence, researchers should conduct validation steps that test for how the measures corresponds to independent information or criteria. Thus, information outside the scope of the textual data in which the measure was constructed serves as an external benchmark (hence "external" evidence).

| ID | Validation Step | Documentation | Considerations | Performance Criteria | Source / References |
|---|---|---|---|---|---|
| Construct Definition and Operationalization | | | | | |
| III.1 | Comparison of measures with human-annotated test set ("gold-standard data") | • | Have I reached sufficient predictive performance on a test set of held-out human annotations? Did I apply cross-validation to calculate average performance metrics? | Evaluating performance metrics (i.e., F1) for dataset of human annotations | (Samory et al., 2021) |
| Surrogate Label Comparison (context-specific) | | | | | |
| V.5 | Comparison of measures with surrogate labels | • | Have I reached sufficient predictive performance on the surrogate labels? | Evaluating performance metrics (i.e., F1) for the surrogate labels | Grimmer et al. (2022) |
| Criterion Prediction (context-specific) | | | | | |
| V.6 | Criterion Prediction | • | Have I been able to accurately predict real-word phenomena? | Evaluating predictive metrics (i.e., regression coefficient) for the criteria | Grimmer et al. (2022) |

## Robustness Checks

Next to the three types of validation evidence outlined above, the ValiText framework also recommends the continuous test of robustness checks to assess the impact of researchers' degree of freedom on the measurement outcome. On a general note, one could see robustness checks as additional means to test whether decisions regarding the measurement design might have a sustainable effect on the measure's outcome.

| ID | Validation Step | Documentation | Considerations | Performance Criteria | Source / References |
|---|---|---|---|---|---|
| Construct Definition and Operationalization | | | | | |
| IV.1 | Rerunning the analysis using alternative text models | • | Have I rerun the analysis with alternative text-based methods, such as a baseline model? | Displaying performance metrics (e.g., F1 score on human annotated test set (see III.1)) for alternative measurements | (Samory et al., 2021) |
| IV.2 | Rerunning the analysis using different hyperparameter settings | • | Have I rerun the analysis with alternative hyperparameter settings? | Displaying performance metrics (e.g., F1 score on human annotated test set (see III.1)) for alternative hyperparameter settings | (Arnold et al., 2023) |
| IV.3 | Rerunning the analysis using different cutoff thresholds | • | Have I rerun the analysis with alternative cutoff thresholds? | Displaying performance metrics (e.g., F1 score on human annotated test set (see III.1)) | Grimmer et al. (2022) |
| | | | | for alternative cut-off thresholds | |
| IV.4 | Rerunning the analysis using different text cleaning and preprocessing steps | • | Have I rerun the analysis using alternative data cleaning and preprocessing settings (e.g., removing certain phrases or features of the data)? | Displaying performance metrics (e.g., F1 score on human annotated test set (see III.1)) for alternative preprocessing steps | Grimmer et al. (2022) |